# Decompose and Leverage Preferences from Expert Models for Improving Trustworthiness of MLLMs

Rui Cao[†]  Yuming Jiang[*]  Michael Schlichtkrull[‡]  Andreas Vlachos[†]

[†]University of Cambridge  [*]Nanyang Technological University  [‡]Queen Mary University of London

{rc990,av308}@cam.ac.uk  yuming002@e.ntu.edu.sg  m.schlichtkrull@qmul.ac.uk

## Abstract

*Multimodal Large Language Models (MLLMs) can enhance trustworthiness by aligning with human preferences. As human preference labeling is laborious, recent works employ evaluation models for assessing MLLMs' responses, using the model-based assessments to automate preference dataset construction. This approach, however, faces challenges with MLLMs' lengthy and compositional responses, which often require diverse reasoning skills that a single evaluation model may not fully possess. Additionally, most existing methods rely on closed-source models as evaluators. To address limitations, we propose* **DecompGen**, *a decomposable framework that uses an ensemble of open-sourced expert models. DecompGen breaks down each response into atomic verification tasks, assigning each task to an appropriate expert model to generate fine-grained assessments. The DecompGen feedback is used to automatically construct our preference dataset, DGPref. MLLMs aligned with DGPref via preference learning show improvements in trustworthiness, demonstrating the effectiveness of DecompGen*[1].

## 1. Introduction

Leveraging the success of Large Language Models (LLMs), recent works have attempted to extend LLMs to the vision-language setting by learning to align visual features to the text space of LLMs, resulting in Multimodal Large Language Models (MLLMs) [10, 20, 24, 25, 45]. Though achieving remarkable performance in several vision-language tasks [8, 25, 26], MLLMs often exhibit overconfidence and provide problematic responses, such as generating descriptions that inaccurately reflect image content content [16, 34, 40, 42], as illustrated in Fig. 1.

Existing research shows that aligning MLLMs with human preferences can significantly enhance trustworthiness of responses anchoring the visual context [40, 47]. However, human preference annotation is expensive, laborious and can be biased [47]. Alternatively, some works adopts heuristic rules for automatic preference data generation [15, 53], which are formulaic and unnatural. Another line of works exploit an evaluation model (Eval-M) as a judge for feedback generation, improving MLLMs by aligning them with preference data based on Eval-M assessments [21, 44, 48, 51]. MLLMs aligned with preference data derived from model-based feedback have demonstrated notable gains in trustworthiness. Despite this progress, several limitations remain in leveraging an Eval-M for preference data collection.

---
[1]Code available: https://github.com/abril4416/DGPref

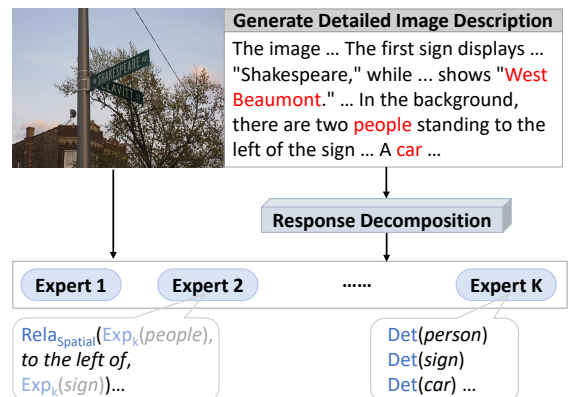

Figure 1. **A problematic response about detailed image descriptions from LLaVA-v1.5.** We proposed **DecompGen**, a decomposable and interpretable framework to assess MLLMs' responses. It decomposes responses to atomic verification tasks, assigns atomic tasks to proper expert models and generate fine-grained feedback with an ensemble of expert models. Expert $K$ stands for the k-th expert model.

Firstly, responses from MLLMs (even at the sentence-level) are compositional. For instance, checking the sentence "*two people standing to the left of the sign*" in Fig. 1 entangles object detection, counting and spatial reasoning. However, Eval-Ms struggle with compositional rea-soning [9, 29, 41], thus, it is challenging for an Eval-M to assess and provide accurate feedback. Secondly, an Eval-M may not be not all-rounder. For example, Eval-Ms are poor at spatial reasoning [17, 41]. Therefore, generating feedback for compositional MLLM responses with a single Eval-M could be inaccurate, leading to failure of further preference dataset construction based on the feedback. Thirdly, existing works predominantly rely on a powerful but closed-source Eval-M (e.g., GPT-4 [31]) for feedback generation [21, 44, 51], which is expensive.

To solve the limitations above, we introduced **DecompGen**, a decomposable framework, using an ensemble of open-source expert models. DecompGen consists of two steps: response decomposition and expert model execution. Specifically, given a response, DecompGen generates a response-specific layout by decomposing it into atomic verification tasks (e.g., object existence verification and spatial relation verification). Then, expert models will be dynamically assembled and executed according to the layout, where each model will be responsible for a specific atomic verification task. Figure 1 illustrates a part of the expert executions. Fine-grained feedback from the execution of expert models will be aggregated for constructing our proposed preference dataset, **DGPref**.

We apply Direct Preference Optimization (DPO) [33] to align MLLMs with DGPref. Experimental results demonstrate the trustworthiness of MLLMs is largely enhanced after preference learning with DGPref. Moreover, MLLMs aligned with DGPref surpass MLLMs aligned with other preference datasets. We summarize our contributions as follows:

- We present DecompGen, a decomposable framework using an ensemble of experts for generating high-quality feedback to MLLMs' responses.
- The DecompGen generated feedback is then used for automatically constructing a preference dataset, DGPref. MLLMs after preference alignment with DGPref enhance trustworthiness significantly.
- MLLMs aligned with DGPref surpass MLLMs aligned with other preference datasets, confirming the effectiveness of DecompGen, which leads to a high-quality DGPref. Extensive ablations and analysis are conducted to better understand the advantages and limitations of our proposed method.

## 2. Method

In this section, we elaborate on the proposed **DecompGen** for generating fine-grained and high-quality feedback to MLLMs' responses. It firstly decomposes a response into atomic verification tasks (Sec. 2.1). Based on decomposition results, expert models, each excel in a specific reasoning task, will be assembled to generate feedback (Sec. 2.2). The DecompGen feedback will be used for constructing a preference dataset, **DGPref** (Sec. 2.3), which will then be exploited for preference optimization of MLLMs (Sec. 2.4).

### 2.1. Response Decomposition

Responses from MLLMs could be lengthy. For instance, the average length of detailed image descriptions from LLaVA-1.5 [24] exceeds 100 words. Moreover, not all parts of a response are directly related to the visual context. Some elaborations, such as "*The atmosphere is warm and festive*" or "*It would be great to enjoy holiday season with loved ones.*", are subjective and not directly depicted in the image. Considering this, we introduce a two-step decomposition for model responses. In the first step, a lengthy response will be decomposed into short check-worthy parts related to multiple visually check-worthy aspects. In the following step, each check-worthy part will be further decomposed into atomic verification tasks.

For the definition of "visually check-worthy", we follow previous works [7, 50] and consider *object existence*, *object relations*, *object attributes*, *counts of objects* and *image texts* as visually check-worthy. To decompose a response into short and check-worthy parts, we exploit the in-context learning capability of LLMs. Specifically, for each aspect (e.g., object existence), we provide eight in-context examples and prompt an LLM to extract related parts from a response. The exact prompting templates used for visually check-worthy part extraction are available in Appendix N. For object existence, object entities mentioned in a response are extracted. For object relations, relevant parts are extracted and converted into triplets consisting of the subject, the relation and the object (e.g., *(people, to the left of, sign)*). Relevant parts about object attributes are extracted and represented with tuples consisting of the attribute and the associated object (e.g., *(standing, people)*). For object counts, the number and the associated object (e.g., *(two, people)*) are extracted. For image texts, we extract parts in a response related to describing texts presented in an image. An example of check-worthy part extraction is provided in Fig. 2.

We further decompose the verification of extracted check-worthy parts into atomic reasoning tasks, inspired by previous work [2, 5, 12]. For parts about object existence, we treat it as an object detection task (`[DET]`). For parts related to object relations, a hierarchical verification will be conducted: validation for the existence of both the subject and the object will be performed firstly, followed by the verification of their relationship (`[RELA]`). Similarly, for object attributes, we first check the existence of the object in the extracted part and perform attribute verification of the object (`[ATTR]`). Object counts are checked by the detection of the object and then validating the equality between the number of detected instances and the extracted number (`[COUNT]`). For parts associated with image texts, we regard the verification as optical character recognition task ([OCR]). [·] denotes an atomic task. Details of how verification of extracted parts about different check-worthy aspects is decomposed into atomic verification tasks is illustrated in Table 1.

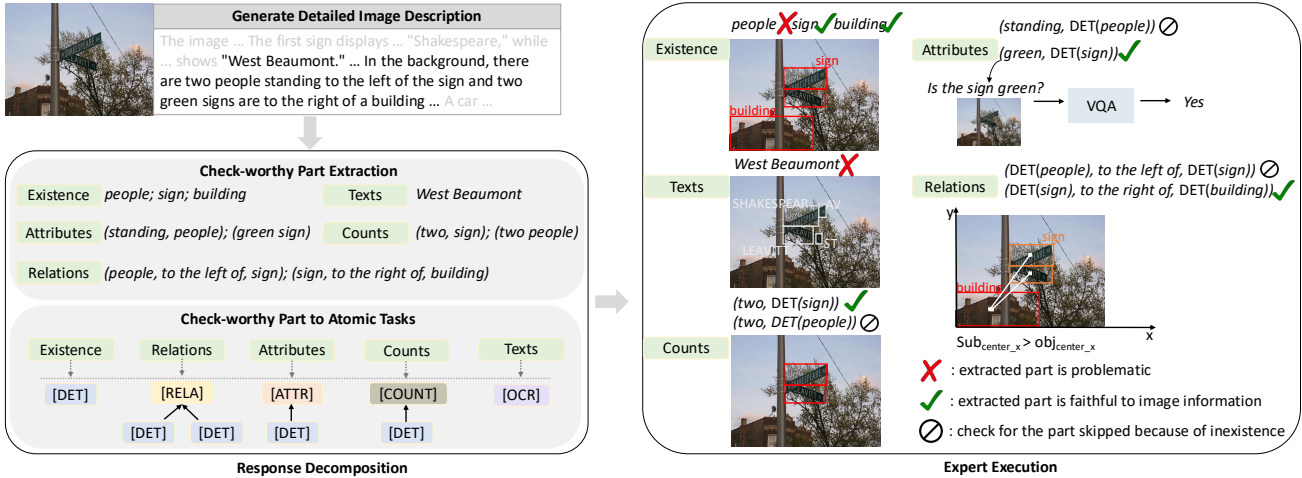

Figure 2. **Overview of the DecompGen model for feedback generation.** It extracts visually check-worthy parts of a model response, decomposes each part into atomic verification tasks and assembles and executes expert models for fine-grained feedback generation.

| C-W Asp. | Atomic Tasks |
|---|---|
| Obj. Existence | [DET](*obj*) |
| Obj. Relation | [RELA]([DET](*obj*$_s$), *rela*, [DET](*obj*$_o$)) |
| Obj. Attribute | [ATTR](*attr*, [DET](*obj*)) |
| Counts of Obj. | [COUNT](*num*, [DET](*obj*)) |
| Image Text | [OCR](*text*) |

Table 1. **The mapping from the verification of extracted parts about different check-worthy aspects (C-W Asp.) to atomic verification tasks.** *obj*, *rela*, *attr*, *num* and *text* are from extracted contents and are short for object, relation, attribute, number and image text, respectively.

### 2.2. Expert Execution

After the decomposition of the response into atomic verification tasks, expert models, each excel in an atomic task, will be assembled and executed. The expert executions will lead to an assessment for each check-worthy part, as illustrated in Fig. 2. Here we use a score to denote the assessment, where 0 means the check-worthy part is faithful to the image content while $-1$ denotes the check-worthy part is problematic (i.e., not align with the image). Below, we elaborate how experts are executed for parts related to different check-worthy aspects:

**Object Existence.** The expert for [DET] is called to detect the object *obj*. If the expert model returns no bounding boxes (i.e., cannot detect the object), a feedback score of $-1$ is assigned. If the object is detected, the part receives a score of 0. We exploit an open-vocabulary object detection model [28] for [DET].

**Image Text.** For extracted image texts mentioned in a response, we match it with detected image scene texts from the expert for [OCR]. If an exact match could be found, we score the piece of image text as 0, otherwise, $-1$. An OCR detection tool[2] is employed for [OCR].

**Object Relation.** Given an extracted relation triplet about object relations, the process begins with running the [DET] expert to confirm the existence of the subject ($obj_s$) and the object ($obj_o$). If either cannot be detected, relation verification is skipped, as the existence of both entities is a prerequisite for evaluating their relationship. Once the existence of both entities is confirmed, the expert for [RELA] will be executed for relation verification. If the expert affirms the relationship, a score of 0 is assigned. Otherwise, a score of $-1$ is given. We separate [RELA] into general relation verification and spatial relation verification, where general relations are verified with a VQA model [20] while spatial relations are verified with heuristic rules over coordinates of detected bounding boxes for the subject and the object.

**Object Attribute.** Feedback generation for parts about object attributes begins with the [DET] expert. If the object is not detected, further verification is skipped. If bounding boxes are detected, the [ATTR] expert is called for attribute verification. If the attribute is confirmed, the part receives a score of 0; otherwise, it is assigned a score of $-1$. [ATTR] is also divided into general attributes and size-related attributes, where general attribute verification is assigned to a VQA model [20] while size-related attributes are verified with heuristic rules over the height and the width of the detected instance.

---
[2]https://github.com/JaidedAI/EasyOCR

**Counts of Objects.** The expert model for [COUNT] will be executed after the confirmation of the object existence. [COUNT] is implemented by heuristics rules, checking if the number of bounding boxes returned by [DET] matches the number of counts in the extracted part. If the numbers differ, the part is given a score of $-1$, while a match results in a score of $0$.

The choices of expert models, definition of heuristics and details about how expert models perform atomic tasks are provided in Appendix A. It should be noted that all expert models are replaceable. After expert executions, we have fine-grained feedback scores for all check-worthy parts, which will be utilized for preference dataset construction.

### 2.3. Preference Data Generation

Generally, given an image $\mathcal{I}$ and an associated query $\mathcal{Q}$, a preference dataset provides annotations to categorize MLLMs' responses into preferred ones ($\mathcal{R}_{\text{pref}}$) and rejected ones ($\mathcal{R}_{\text{rej}}$). Here we focus on the type of query about detailed image descriptions as existing studies found MLLMs are prone to provide problematic responses to this query type [22]. Therefore, two steps are needed for preference data collection: 1) diverse response generation for detailed image descriptions; 2) categorizing responses into preferred ones and rejected ones.

Firstly, to collect diverse responses about detailed image captioning, we prompt MLLMs with eight different instructions, with diverse semantics, all about detailed image description generation following previous work [47] and add them to a response pool. The instructions we use are provided in Appendix N. Secondly, we use DecompGen to generate fine-grained feedback scores to each response, as introduced in Sec. 2.2. Weighted averaging is applied over the fine-grained feedback scores to generate an overall score to a response. Then, we generate preference data using pairwise combinations of responses from the response pool. The response with a higher overall score will be regarded as $\mathcal{R}_{\text{pref}}$, while the other with lower score as $\mathcal{R}_{\text{rej}}$. With the two steps above, we use DecompGen to construct a preference dataset, **DGPref** (**D**ecompGen Feedback **G**enerated **Pref**erance dataset).

### 2.4. Model Alignment

Based on the DGPref, we choose to use Direct Preference Optimization (DPO) [33] for preference learning. DPO is motivated by the reward modeling in Reinforcement Learning from Human Feedback (RLHF) [32] and tries to maximize the difference of rewards between choosing the preferred response and rejected response. Compared to traditional RLHF, DPO is more stable and computationally lightweight and is now a popular choice for preference optimization in multimodal scenarios [21, 48, 51].

Given an input instruction $\mathcal{Q}$ and an image $\mathcal{I}$, the MLLM provides a response $\mathcal{R}$. DPO formulates the reward $r$ as:

$$r(\mathcal{Q}, \mathcal{I}, \mathcal{R}) = \beta \log \frac{\pi_\theta(\mathcal{R}|\mathcal{Q}, \mathcal{I})}{\pi_{\text{ref}}(\mathcal{R}|\mathcal{Q}, \mathcal{I})} + Z(\mathcal{Q}, \mathcal{I}), \quad (1)$$

where $\pi_\theta$ denotes the model parameters, $\pi_{\text{ref}}$ is the parameters of the base model without DPO tuning, $Z(\cdot)$ is the partition function and $\beta$ is a hyper-parameter, controlling the deviation from base model. The preference learning is converted to maximize the difference of rewards between preferred and rejected data points [4]:

$$L_\theta = -\beta(\log \sigma(\log \frac{\pi_\theta(\mathcal{R}_{\text{pref}}|\mathcal{Q}, \mathcal{I})}{\pi_{\text{ref}}(\mathcal{R}_{\text{pref}}|\mathcal{Q}, \mathcal{I})} - \log \frac{\pi_\theta(\mathcal{R}_{\text{rej}}|\mathcal{Q}, \mathcal{I})}{\pi_{\text{ref}}(\mathcal{R}_{\text{rej}}|\mathcal{Q}, \mathcal{I})})$$
$$\approx \sigma(r(\mathcal{Q}, \mathcal{I}, \mathcal{R}_{\text{pref}}) - r(\mathcal{Q}, \mathcal{I}, \mathcal{R}_{\text{rej}})), \quad (2)$$

where $\sigma$ is the Sigmoid function. The model parameters will be updated according to Equation 2 for preference optimization with DGPref.

## 3. Experiments

### 3.1. Evaluation Setting

**Evaluation Datasets.** We evaluate hallucinations of MLLMs on three benchmarks: (1) **ObjHal** [34] evaluates object hallucinations in detailed image descriptions generated by MLLMs with a rule-based evaluation mechanism, (2) **MMHal** [40] leverages GPT-4 [31] to compare models' responses and human responses to assess the overall informativeness and hallucination rate. (3) **AMBER** [42] focuses on multiple dimensions of hallucinations, including existence, attribute and relation hallucinations. It measures both the hallucination rate and coverage of responses. Statistics of evaluation datasets are available in Appendix I.

**Evaluation Metrics.** We follow official evaluation protocols of these datasets [34, 40, 42]. For ObjHal, we report both the response-level hallucination rate (CHAIR$_s$) and the instance-level hallucination rate (CHAIR$_i$). For MMHal, we report informative score (0 - 6) using GPT-4 to compare the model's response with the human annotated response, and a hallucination rate (HalRate) by comparing the model's response with fine-grained object annotations. For AMBER, four metrics were used for evaluation. Specifically, softer versions of instance-level hallucination rate (s.CHAIR$_i$) and response-level hallucination rate (HalRate), considering the semantic similarity between predictions and annotations, are included. In addition, the coverage of a response (COVER.) and the similarity between model hallucinations and human cognition (Cog.) are reported.

**Implementation Details.** To validate the proposed method, we use two MLLMs as the base model, LLaVA-v1.5 [24] and Qwen-VL-Chat [3], both with 7B model parameters. Visual Genome (VG) [18] is the image source for the construction of DGPref, which resulted in 16k images with 52k preference data samples. For response decomposition, we use the Llama-3.1-8B-Instruct model [1]. During DPO training, instead of updating all model parameters, we apply parameter-efficient tuning technique, low-rank adaptation (LoRA) [11]. To avoid randomness in testing, we set the sampling temperature to 0 [48]. The max length of generation is set to be 1024. More details about the implementation are available in Appendix I.

| Model | ObjHal | | MMHal | | AMBER | | | |
|---|---|---|---|---|---|---|---|---|
| | CHAIR$_s$ ↓ | CHAIR$_i$ ↓ | Score ↑ | HalRate ↓ | s.CHAIR$_i$ ↓ | COVER. ↑ | HalRate ↓ | Cog. ↓ |
| GPT-4V | 13.6 | 7.3 | 3.49 | 0.28 | 4.6 | 67.1 | 30.7 | 2.6 |
| *Error Reduction with Decoding Strategies* | | | | | | | | |
| VCD | 48.8 | 24.3 | 2.12 | 0.54 | - | - | - | - |
| Less-is-more | 40.3 | 17.8 | 2.33 | 0.50 | 5.1 | 49.1 | 22.7 | 2.0 |
| OPERA | 45.1 | 22.3 | 2.15 | 0.54 | - | - | - | - |
| LURE | 27.7 | 17.3 | 1.64 | 0.60 | - | - | - | - |
| *Error Reduction with Preference Data* | | | | | | | | |
| RLHF$^\dagger$ | 38.1 | 18.9 | 2.02 | 0.63 | 7.7 | 52.1 | 39.0 | 4.4 |
| RLHF-V$^{\dagger*}$ | 12.2 | 7.5 | 2.81 | 0.49 | 6.3 | 46.1 | 25.1 | 2.1 |
| HACL$^*$ | - | - | 2.13 | 0.50 | - | - | - | - |
| POVID | 48.1 | 24.4 | 2.08 | 0.56 | - | - | - | - |
| HALVA | 41.4 | 11.7 | 2.25 | 0.54 | 6.6 | 53.0 | 32.2 | 2.4 |
| HA-DPO | 39.9 | 19.9 | 1.97 | 0.60 | 6.7 | 49.8 | 30.9 | 3.3 |
| Silkie | 25.3 | 13.9 | 3.01 | 0.41 | 5.4 | **55.8** | 29.0 | 2.0 |
| HSA-DPO | 11.0 | 5.5 | <u>3.07</u> | <u>0.34</u> | 3.7 | 52.4 | 19.0 | 1.6 |
| RLAIF-V$^*$ | <u>8.5</u> | 4.3 | 3.06 | **0.29** | 3.1 | 50.7 | 16.3 | 1.0 |
| *Base MLLMs* | | | | | | | | |
| LLaVA-v1.5 | 54.7 | 15.9 | 2.19 | 0.57 | 7.4 | 51.8 | 34.7 | 4.1 |
| Qwen | 36.0 | 21.3 | 2.89 | 0.43 | 6.6 | 53.2 | 31.2 | 2.9 |
| *Base MLLMs with DGPref* | | | | | | | | |
| DGPref$_{\text{LLaVA}}$ | 10.3 | <u>2.6</u> | 2.59 | 0.36 | **1.2** | 51.2 | **7.8** | **0.5** |
| DGPref$_{\text{Qwen}}$ | **8.0** | **2.1** | **3.19** | 0.34 | <u>1.5</u> | 54.0 | 10.0 | <u>0.8</u> |

Table 2. **Performance of models on evaluation benchmarks.** The best and the second best results of open-source MLLMs are shown in **bold** and <u>underline</u>, respectively. We use $^\dagger$ to denote models larger than 7B, as they do not have results with 7B MLLMs (the performance and analysis of baselines with larger model sizes is available in Appendix D). We use $^*$ to denote MLLMs with full fine-tuning during preference learning.

**Baselines.** The first line of works reduces errors with carefully designed decoding strategies (e.g., post-hoc processing or probability manipulations during decoding), without model tuning. VCD [19], Less-is-more [49], OPERA [13] and LURE [54] all fall into this category. The second line of works tries to mitigate errors by learning from preference data. Some collect preference data via human annotation (e.g., RLHF [40] and RLHF-V [47]), some collect preference data with heuristic rules (e.g., HACL [15], POVID [53] and HALVA [35]) and some leverage an evaluation model for assessing responses to construct preference datasets (e.g., HA-DPO [51], Silkie [21], HSA-DPO [44] and RLAIF-V [48]). We also include the performance of GPT-4 [31] for a reference to demonstrate to what extent our proposed method can bridge the gap between open-source MLLMs and more powerful closed-source MLLMs.

### 3.2. Experimental Results

**Comparison with Base MLLMs.** We compare the base MLLMs, LLaVA-v1.5 [24] and Qwen-VL-Chat [3], before and after preference learning with DGPref (denoted as DGPref$_{\text{LLaVA}}$ and DGPref$_{\text{Qwen}}$). By comparing the performance shown in the last two blocks in Table 2, we observe obvious drops in hallucination rates after preference alignment. Besides, we also observe that MLLMs after alignment achieve comparable or better informativeness (e.g., an increased Score on MMHal and a comparbale or higher COVER. on AMBER). It shows that the reduction in hallu-cinations is not at the sacrifice of informativeness.

| Model | ObjHal | | MMHal | | AMBER | | | |
|---|---|---|---|---|---|---|---|---|
| | CHAIR$_s$ ↓ | CHAIR$_i$ ↓ | Score ↑ | HalRate ↓ | s.CHAIR$_i$ ↓ | COVER. ↑ | HalRate ↓ | Cog. ↓ |
| Base | 36.0 | 21.3 | 2.89 | 0.43 | 6.6 | 53.2 | 31.2 | 2.9 |
| Pref$_{Obj}$ | 8.7 | 2.1 | 2.65 | 0.45 | 1.5 | 53.9 | 9.9 | 0.6 |
| Pref$_{Obj-0.1}$ | 15.0 | 3.5 | 2.79 | 0.49 | 2.1 | 53.1 | 13.6 | 0.7 |
| Pref$_{Obj-GT}$ | 1.7 | 1.0 | 2.51 | 0.55 | 2.9 | 46.1 | 13.9 | 0.6 |
| DGPref$_{COCO}$ | 11.3 | 2.8 | 3.11 | 0.37 | 1.9 | 50.3 | 12.7 | 1.2 |
| DGPref$_{half}$ | 13.0 | 3.2 | 3.12 | 0.40 | 1.9 | 54.3 | 13.0 | 0.9 |
| DGPref | 8.0 | 2.1 | 3.19 | 0.34 | 1.5 | 54.0 | 10.1 | 0.8 |

Table 3. **Ablation results of DecompGen for preference dataset construction, based on Qwen-VL-Chat.** We find out that 1) considering all five check-worthy aspect as mentioned in Sec. 2.1, 2) individual expert performance and 3) source of images are important to DecompGen for preference data collection. The amount of collected preference data plays a vital role in MLLMs' preference learning.

**Comparison with Baselines.** In Table 2, we compare the proposed method with existing works for enhancing trustworthiness of MLLMs. We find methods reducing hallucination with the help of preference data generally achieve better performance than methods reducing errors with decoding strategies, highlighting the effectiveness of preference data.

Comparing with the line of works reducing errors using preference data, DGPref$_{Qwen}$ achieves lower or comparable error rates even in comparison with the strongest baselines (e.g., Silkie, HAS-DPO and RLAIF-V). Besides, DGPref$_{Qwen}$ has a good balance between hallucination mitigation and informativeness (i.e., a good coverage of image content in detailed descriptions, shown with COVER. on AMBER) compared to other baselines. It demonstrates the superiority our DGPref and DecompGen over other preference data-based methods.

Besides, the proposed DecompGen is also more efficient. Both Silkie and HSA-DPO use GPT-4 as the evaluation model for preference dataset construction and RLAIF-V leverages a 34B model. In contrast, the ensemble of expert models in DecompGen only has 4B parameters. By decomposing the complex, compositional response assessing task into atomic tasks, DecompGen can generate more accurate assessments with smaller models.

Fianally, after preference alignment, DGPref$_{Qwen}$ outperforms GPT-4 on ObjHal and some metrics on AMBER, a promising result given that GPT-4 is speculated to have more than one trillion parameters. We have also shown, while reducing hallucinations of MLLMs, our method has no negative impacts to MLLMs on their standard and finetuned VQA settings (details available in Appendix H).

### 3.3. Ablation Study

**Considered Aspects.** Firstly, we analyze the impact of check-worthy aspects considered in DecompGen. As mentioned in Sec. 2.1, we considered five check-worthy aspects (i.e., object existence, object relations, object attributes, counts of objects and image texts). In this section, we compare preference data collected covering all five aspects (**DGPref**) to data only considering object existence (denoted as **Pref$_{Obj}$**). By comparing the MLLM performance after aligned with Pref$_{Obj}$ and DGPref in Table 3, we observe preference data covering all aspects (i.e., DGPref) leads better trustworthiness of MLLMs, especially on MMHal, as MMHal extensively covers questions about attributes, counting and relations. Therefore, considering all five aspects in DecompGen is essential.

**Impact of Expert Performance.** Since the performance of expert models affects the quality of feedback, in this part, we compare variants of the experts in DecompGen to analyze to what extent the expert performance affects the quality of constructed preference datasets. We take the expert model for object detection task as an example. We vary the performance of the object detection expert in DecompGen by 1) using a lower detection confidence threshold (0.1, compared to 0.25 set originally) and 2) using GT annotations. The preference datasets generated with the two variants of DecompGen are denoted as **Pref$_{Obj-0.1}$** and **Pref$_{Obj-GT}$**, respectively. According to the results in Table 3, aligning MLLMs with Pref$_{Obj-0.1}$, though shows hallucination mitigation on three benchmarks, the quality of which is not as good as that constructed with detection threshold as 0.25 (i.e., Pref$_{Obj}$). The detection expert with a lower detection threshold will be prone to falsely detect non-existent objects, contributing to inaccurate feedback. For instance, even there are hallucinated objects in a response, the detection expert with a threshold 0.1 may not be able to recognize so that Pref$_{Obj-0.1}$ could potentially categorize the response as a preferred response. The low quality of generated preference data incurs failures of preference learning. Good performance of individual experts ensures the quality of DecompGen feedback, contributing to a high quality preference dataset.

The MLLM aligned with Pref$_{Obj-GT}$, as shown in the last row of the second block in Table 3, provides the best performance on ObjHal, which also leverages GT object annotations for evaluation. However, the aligned MLLM is incapable to generalize to MMHal and AMBER. This is likely due to the fact that human annotations consider a limited number of object categories (80 categories), which reduce the diversity of MLLMs' responses. Also, object annotations, although denser compared to annotations for attributes and relations, can hardly cover all object information in an image. Thus aligning with Pref$_{\text{Obj-GT}}$ may falsely penalize a response with objects that exist but not captured by manual object annotations. These reasons lead to pitfalls when using manual annotations to replace the detection expert in DecompGen for preference dataset construction.

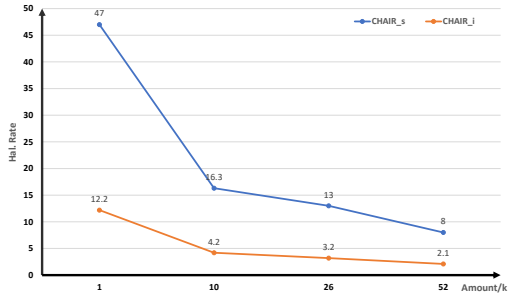

Figure 3. **Performance of Qwen-VL-Chat on ObjHal after preference learning with different amount of data.** Hallucination rates drops with the increase of data amount.

**Image Source.** When constructing DGPref, we leverage Visual Genome (VG) [18] as the image source [35, 44]. In this part, we use COCO [23] as an alternative image source to generate preference data (**DGPref$_{\text{COCO}}$**). The results in Table 3 show that using either DGPref$_{\text{COCO}}$ or DGPref from VG can enhance MLLMs' trustworthiness significantly. We observe using DGPref with VG images can lead to better model performance for preference learning than using DGPref$_{\text{COCO}}$. It could be that VG images have richer content [18] (e.g., diverse objects and dense object relations), which provide richer learning signals for improving MLLMs' trustworthiness.

**Amount of Training Data.** We also analyze the influence of the scale of preference data to preference alignment. We try using half of DGPref (26k) for aligning MLLMs (**DGPref$_{\text{half}}$**). Comparing results in the last two rows of Table 3, the MLLM aligned with the full DGPref generally outperforms that aligned with DGPref$_{\text{half}}$. It proves the scale of the preference data matters in preference learning. Whereas, manual annotation for feedback generation is limited by scale (RLHF-V [47] provides 1.4k data and RLHF [40] provides 10k data). Performance of MLLMs aligned with these scales of preference data (i.e., 1k and 10k) on ObjHal is shown in Fig. 3, which is less competitive. It further highlights the value of automatically constructing high quality preference dataset with DecompGen.

## 3.4. Qualitative Study

**Visualization of Generated Feedback.** We provide visualization of DecompGen feedback in Fig. 4, the preciseness of which is the foundation for a high quality preference dataset. Instead of directly providing overall assessments to long and compositional MLLMs' responses, DecompGen provides interpretable and fine-grained feedback. The decomposable structure of DecompGen also makes it flexible to incorporate any other check-worthy aspects (e.g., commonsense). Meanwhile, it is also easy to pinpoint the weakness of DecompGen by examining outputs from expert models. With the development of foundation models (e.g., object detection models), DecompGen can be facilitated by replacing a weaker expert model with a stronger one, leading to better preference datasets.

**Error Analysis of Generated Feedback.** As described in Sec. 2, DecompGen consists of two stages: response decomposition and expert execution. In this part, we analyze failures of DecompGen due to errors from either stage. Firstly, we manually checked the first 30 examples about their decomposition results. Extracting invalid check-worthy parts (e.g., extracting *other* as an object entity or extracting *wearing ski* as an object attribute) is the most common error type. This type of errors counts for 77% of all errors. There are also few other error types, such as incomplete extraction (e.g., the entity is *restaurant sign* while the LLM extracts *restaurant*) and hallucinations in extraction (i.e., "extract" a part which is not in the model response). We do observe DecompGen misses some check-worthy parts (e.g., it does not extract (*person*, *left*, *giraffe*) of the first example in Fig. 4). However, the missing rate is low (less than 10%). Visualization of commonly seen error cases in response decomposition is available in Appendix L. For errors related to response decomposition, using a stronger LLM can largely solve them according to our preliminary observations.

Next, we analyze errors from expert models. We notice the object detection expert performs well and it can detect objects, which are not predominant in images and are even hard for human to recognize (e.g., *person* in the first example in Fig. 4). Most errors of detection are related to profession entities (e.g., detecting *player* or *pitcher*). Errors in expert execution are more frequently related to general attribute and general relation verification. A plausible reason could be unnatural input questions which are generated with fixed templates. A potential solution is to leverage language models to paraphrase a question given the extracted check-worthy part, rather than using templates for question generation.

**Qualitative Analysis of MLLMs' Responses.** In Table 2, we demonstrate MLLMs can enhance trustworthiness after preference learning with DGPref using automatic evaluation metrics. Here, we perform human evaluation to qual-itatively analyze the effectiveness of preference learning with DGPref. Specifically, we shuffle the three evaluation datasets and randomly select 10 cases from each dataset for human analysis. For each case, we provide annotators, the image, the image-related instruction, the base model response (**Base.**) and the updated model response (**+DGPref**). We ask evaluators to choose which response has fewer errors, is more informative and which response they prefer. Four researchers have participated in the human evaluation part. The human evaluation results are shown in Table 4. According to the results, we notice that the base model aligned with DGPref provides informative and preferable responses with fewer errors most of the time, compared to the base MLLM. It also proves the enhancement in trustworthiness is not at the sacrifice of informativeness. Details for the human evaluation setting and annotator agreements are provided in Appendix M. We also visualize a few responses from the base model and the base model updated with DGPref. The visualization results are shown in Appendix J.

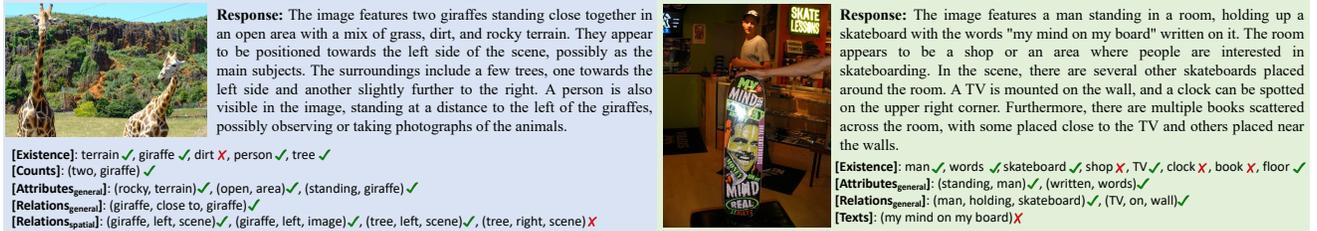

Figure 4. **Visualization of responses and generated feedback.** Skipped verifications are not shown in the figure.

| Model | Fewer Errors | More Inform. | Prefer |
|---|---|---|---|
| **Base.** | 18.3% | 16.7% | 11.7% |
| **+DGPref** | 81.7% | 83.3% | 88.3% |

Table 4. **Human evaluation of responses from the base model and the base model updated with DGPref.**

## 4. Related Work

Vision-language understanding has witnessed significant progress by incorporating Large Language Models (LLMs) into models, resulting Multimodal Large Language Models (MLLMs) [20, 24, 25, 45]. However, recent studies have demonstrated the over-confidence of MLLMs when providing problematic responses inconsistent with image contents [16, 22, 34, 40, 42]. Existing works have designed several methods for enhancing trustworthiness of MLLMs. One line of works adjusts decoding strategies, such as post-processing decoded texts and manipulations of token probabilities [13, 19, 49, 54]. The other line of works focuses on aligning MLLMs with human preferences [40, 47]. Given the high cost and labor of human annotations, recent works proposed to automatically generate preference data [21, 35, 44, 48, 51, 53]. Among works about automatic preference data collection, using an evaluation model (Eval-M) to assess MLLMs' responses and constructing preference datasets based on Eval-M feedback work the best [21, 44, 48]. Though progress made, these methods suffer from the lengthy and compositional MLLMs' responses. Inaccurate assessments from the Eval-M result in a defective preference dataset, which may incur failures of preference learning.

Decomposition has been shown effective towards complex tasks [5, 6, 27, 36, 37, 43]. Motivated by the idea of decomposition, we propose a decomposable framework, DecompGen, which decomposes a complex MLLM response into atomic verification tasks. As a single Eval-M may not be good at all tasks, we leverage a set of expert models, each responsible for an atomic task. The DecompGen assessment from expert models will be applied for automatic preference dataset construction.

## 5. Conclusion

In this work, we propose a decomposable framework, DecompGen, with an ensemble of open-source expert models to provide fine-grained and precise assessments of MLLMs' responses. The DecompGen feedback is then used to automatically construct a preference dataset, DGPref. MLLMs aligned with DGPref demonstrate marked enhancement in trustworthiness. Experimental results confirm the effectiveness, efficiency and interpretability of DecompGen in automating preference data collection.

In this work, though fine-grained assessments provided by DecompGen, we currently rely on binary signals (i.e., preferred and rejected) for preference learning, limited by the existing preference optimization algorithms. With the development of advanced algorithms, DecompGen's detailed assessments could provide richer learning signals for more nuanced training. Another direction is that we can apply our decomposable framework in the pre-training stage, enabling it to assess and filter high-quality data for MLLM pre-training.

## Acknowledgement

Rui Cao and Andreas Vlachos are supported the Alan Turing Institute - DSO Labs grant FEVER-IT. Andreas Vlachos is further supported by the European Research Council grant AVEeriTeC.

## References

[1] Meta AI. Llama 3.1 8b instruct. https://huggingface.co/meta-llama/Llama-3.1-8B-Instruct, 2024. Accessed: 2024-11-14. 5

[2] Jacob Andreas, Marcus Rohrbach, Trevor Darrell, and Dan Klein. Neural module networks. In *2016 IEEE Conference on Computer Vision and Pattern Recognition, CVPR*, pages 39–48, 2016. 2

[3] Jinze Bai, Shuai Bai, Shusheng Yang, Shijie Wang, Sinan Tan, Peng Wang, Junyang Lin, Chang Zhou, and Jingren Zhou. Qwen-vl: A frontier large vision-language model with versatile abilities. *CoRR*, abs/2308.12966, 2023. 4, 5, 14

[4] Ralph Allan Bradley and Milton E Terry. Rank analysis of incomplete block designs: I. the method of paired comparisons. *Biometrika*, 39(3/4):324–345, 1952. 4

[5] Rui Cao and Jing Jiang. Modularized zero-shot VQA with pre-trained models. In *Findings of the Association for Computational Linguistics: ACL*, pages 58–76, 2023. 2, 8, 12

[6] Jifan Chen, Grace Kim, Aniruddh Sriram, Greg Durrett, and Eunsol Choi. Complex claim verification with evidence retrieved in the wild. In *Proceedings of the 2024 Conference of the North American Chapter of the Association for Computational Linguistics: Human Language Technologies (Volume 1: Long Papers), NAACL*, pages 3569–3587, 2024. 8

[7] Chaoyou Fu, Peixian Chen, Yunhang Shen, Yulei Qin, Mengdan Zhang, Xu Lin, Zhenyu Qiu, Wei Lin, Jinrui Yang, Xiawu Zheng, Ke Li, Xing Sun, and Rongrong Ji. MME: A comprehensive evaluation benchmark for multimodal large language models. *CoRR*, abs/2306.13394, 2023. 2

[8] Yash Goyal, Tejas Khot, Douglas Summers-Stay, Dhruv Batra, and Devi Parikh. Making the V in VQA matter: Elevating the role of image understanding in visual question answering. In *2017 IEEE Conference on Computer Vision and Pattern Recognition, CVPR*, pages 6325–6334, 2017. 1, 16

[9] Nitzan Bitton Guetta, Yonatan Bitton, Jack Hessel, Ludwig Schmidt, Yuval Elovici, Gabriel Stanovsky, and Roy Schwartz. Breaking common sense: Whoops! A vision-and-language benchmark of synthetic and compositional images. In *IEEE/CVF International Conference on Computer Vision, ICCV*, pages 2616–2627, 2023. 2

[10] Muyang He, Yexin Liu, Boya Wu, Jianhao Yuan, Yueze Wang, Tiejun Huang, and Bo Zhao. Efficient multimodal learning from data-centric perspective. *CoRR*, abs/2402.11530, 2024. 1

[11] Edward J. Hu, Yelong Shen, Phillip Wallis, Zeyuan Allen-Zhu, Yuanzhi Li, Shean Wang, Lu Wang, and Weizhu Chen. Lora: Low-rank adaptation of large language models. In *The Tenth International Conference on Learning Representations, ICLR*, 2022. 5

[12] Ronghang Hu, Jacob Andreas, Marcus Rohrbach, Trevor Darrell, and Kate Saenko. Learning to reason: End-to-end module networks for visual question answering. In *IEEE International Conference on Computer Vision, ICCV*, pages 804–813. IEEE Computer Society, 2017. 2

[13] Qidong Huang, Xiaoyi Dong, Pan Zhang, Bin Wang, Conghui He, Jiaqi Wang, Dahua Lin, Weiming Zhang, and Nenghai Yu. Opera: Alleviating hallucination in multi-modal large language models via over-trust penalty and retrospection-allocation. In *IEEE Conference on Computer Vision and Pattern Recognition, CVPR*, pages 13418–13427, 2024. 5, 8

[14] Drew A. Hudson and Christopher D. Manning. GQA: A new dataset for real-world visual reasoning and compositional question answering. In *IEEE Conference on Computer Vision and Pattern Recognition, CVPR*, pages 6700–6709, 2019. 15

[15] Chaoya Jiang, Haiyang Xu, Mengfan Dong, Jiaxing Chen, Wei Ye, Ming Yan, Qinghao Ye, Ji Zhang, Fei Huang, and Shikun Zhang. Hallucination augmented contrastive learning for multimodal large language model. In *IEEE Conference on Computer Vision and Pattern Recognition, CVPR*, pages 27036–27046, 2024. 1, 5

[16] Chaoya Jiang, Wei Ye, Mengfan Dong, Hongrui Jia, Haiyang Xu, Ming Yan, Ji Zhang, and Shikun Zhang. Hal-eval: A universal and fine-grained hallucination evaluation framework for large vision language models. *CoRR*, abs/2402.15721, 2024. 1, 8

[17] Amita Kamath, Jack Hessel, and Kai-Wei Chang. What's "up" with vision-language models? investigating their struggle with spatial reasoning. In *Proceedings of the 2023 Conference on Empirical Methods in Natural Language Processing, EMNLP*, pages 9161–9175, 2023. 2, 12

[18] Ranjay Krishna, Yuke Zhu, Oliver Groth, Justin Johnson, Kenji Hata, Joshua Kravitz, Stephanie Chen, Yannis Kalantidis, Li-Jia Li, David A. Shamma, Michael S. Bernstein, and Li Fei-Fei. Visual genome: Connecting language and vision using crowdsourced dense image annotations. *Int. J. Comput. Vis.*, 123(1):32–73, 2017. 4, 7

[19] Sicong Leng, Hang Zhang, Guanzheng Chen, Xin Li, Shijian Lu, Chunyan Miao, and Lidong Bing. Mitigating object hallucinations in large vision-language models through visual contrastive decoding. In *IEEE Conference on Computer Vision and Pattern Recognition, CVPR*, pages 13872–13882, 2024. 5, 8

[20] Junnan Li, Dongxu Li, Silvio Savarese, and Steven C. H. Hoi. BLIP-2: bootstrapping language-image pre-training with frozen image encoders and large language models. In *International Conference on Machine Learning, ICML*, pages 19730–19742, 2023. 1, 3, 8, 12

[21] Lei Li, Zhihui Xie, Mukai Li, Shunian Chen, Peiyi Wang, Liang Chen, Yazheng Yang, Benyou Wang, and Lingpeng Kong. Silkie: Preference distillation for large visual language models. *CoRR*, abs/2312.10665, 2023. 1, 2, 4, 5, 8, 16

[22] Yifan Li, Yifan Du, Kun Zhou, Jinpeng Wang, Wayne Xin Zhao, and Ji-Rong Wen. Evaluating object hallucination in

large vision-language models. In *Proceedings of the 2023 Conference on Empirical Methods in Natural Language Processing, EMNLP*, pages 292–305, 2023. 4, 8, 16

[23] Tsung-Yi Lin, Michael Maire, Serge J. Belongie, James Hays, Pietro Perona, Deva Ramanan, Piotr Dollár, and C. Lawrence Zitnick. Microsoft COCO: common objects in context. In *Computer Vision - ECCV*, pages 740–755, 2014. 7

[24] Haotian Liu, Chunyuan Li, Yuheng Li, and Yong Jae Lee. Improved baselines with visual instruction tuning. *CoRR*, abs/2310.03744, 2023. 1, 2, 4, 5, 8, 14, 15, 16

[25] Haotian Liu, Chunyuan Li, Qingyang Wu, and Yong Jae Lee. Visual instruction tuning. In *Advances in Neural Information Processing Systems 36: Annual Conference on Neural Information Processing Systems NeurIPS*, 2023. 1, 8

[26] Kenneth Marino, Mohammad Rastegari, Ali Farhadi, and Roozbeh Mottaghi. OK-VQA: A visual question answering benchmark requiring external knowledge. In *IEEE Conference on Computer Vision and Pattern Recognition, CVPR*, pages 3195–3204, 2019. 1

[27] Sewon Min, Kalpesh Krishna, Xinxi Lyu, Mike Lewis, Wen-tau Yih, Pang Wei Koh, Mohit Iyyer, Luke Zettlemoyer, and Hannaneh Hajishirzi. Factscore: Fine-grained atomic evaluation of factual precision in long form text generation. In *Proceedings of the 2023 Conference on Empirical Methods in Natural Language Processing, EMNLP*, pages 12076–12100, 2023. 8

[28] Matthias Minderer, Alexey A. Gritsenko, Austin Stone, Maxim Neumann, Dirk Weissenborn, Alexey Dosovitskiy, Aravindh Mahendran, Anurag Arnab, Mostafa Dehghani, Zhuoran Shen, Xiao Wang, Xiaohua Zhai, Thomas Kipf, and Neil Houlsby. Simple open-vocabulary object detection with vision transformers. *CoRR*, abs/2205.06230, 2022. 3, 12

[29] Chancharik Mitra, Brandon Huang, Trevor Darrell, and Roei Herzig. Compositional chain-of-thought prompting for large multimodal models. In *IEEE/CVF Conference on Computer Vision and Pattern Recognition, CVPR*, pages 14420–14431, 2024. 2

[30] John X. Morris, Eli Lifland, Jin Yong Yoo, Jake Grigsby, Di Jin, and Yanjun Qi. Textattack: A framework for adversarial attacks, data augmentation, and adversarial training in NLP. In *Proceedings of the 2020 Conference on Empirical Methods in Natural Language Processing: System Demonstrations, EMNLP*, pages 119–126, 2020. 12

[31] OpenAI. GPT-4 technical report. *CoRR*, abs/2303.08774, 2023. 2, 4, 5

[32] Long Ouyang, Jeffrey Wu, Xu Jiang, Diogo Almeida, Carroll L. Wainwright, Pamela Mishkin, Chong Zhang, Sandhini Agarwal, Katarina Slama, Alex Ray, John Schulman, Jacob Hilton, Fraser Kelton, Luke Miller, Maddie Simens, Amanda Askell, Peter Welinder, Paul F. Christiano, Jan Leike, and Ryan Lowe. Training language models to follow instructions with human feedback. In *Advances in Neural Information Processing Systems 35: Annual Conference on Neural Information Processing Systems NIPS*, 2022. 4

[33] Rafael Rafailov, Archit Sharma, Eric Mitchell, Christopher D. Manning, Stefano Ermon, and Chelsea Finn. Direct preference optimization: Your language model is secretly a reward model. In *Advances in Neural Information Processing Systems 36: Annual Conference on Neural Information Processing Systems NeurIPS*, 2023. 2, 4

[34] Anna Rohrbach, Lisa Anne Hendricks, Kaylee Burns, Trevor Darrell, and Kate Saenko. Object hallucination in image captioning. In *Proceedings of the 2018 Conference on Empirical Methods in Natural Language Processing*, pages 4035–4045, 2018. 1, 4, 8

[35] Pritam Sarkar, Sayna Ebrahimi, Ali Etemad, Ahmad Beirami, Sercan Ö. Arik, and Tomas Pfister. Mitigating object hallucination via data augmented contrastive tuning. *CoRR*, abs/2405.18654, 2024. 5, 7, 8

[36] Timo Schick, Jane Dwivedi-Yu, Roberto Dessì, Roberta Raileanu, Maria Lomeli, Eric Hambro, Luke Zettlemoyer, Nicola Cancedda, and Thomas Scialom. Toolformer: Language models can teach themselves to use tools. In *Advances in Neural Information Processing Systems 36: Annual Conference on Neural Information Processing Systems 2023, NeurIPS*, 2023. 8

[37] Michael Schlichtkrull, Zhijiang Guo, and Andreas Vlachos. Averitec: A dataset for real-world claim verification with evidence from the web. In *Advances in Neural Information Processing Systems 36: Annual Conference on Neural Information Processing Systems 2023, NeurIPS*, 2023. 8

[38] Amanpreet Singh, Vivek Natarajan, Meet Shah, Yu Jiang, Xinlei Chen, Dhruv Batra, Devi Parikh, and Marcus Rohrbach. Towards VQA models that can read. In *IEEE Conference on Computer Vision and Pattern Recognition, CVPR*, pages 8317–8326, 2019. 15

[39] Sanjay Subramanian, William Merrill, Trevor Darrell, Matt Gardner, Sameer Singh, and Anna Rohrbach. Reclip: A strong zero-shot baseline for referring expression comprehension. In *Proceedings of the 60th Annual Meeting of the Association for Computational Linguistics (Volume 1: Long Papers), ACL*, pages 5198–5215, 2022. 12

[40] Zhiqing Sun, Sheng Shen, Shengcao Cao, Haotian Liu, Chunyuan Li, Yikang Shen, Chuang Gan, Liangyan Gui, Yu-Xiong Wang, Yiming Yang, Kurt Keutzer, and Trevor Darrell. Aligning large multimodal models with factually augmented RLHF. In *Findings of the Association for Computational Linguistics, ACL*, pages 13088–13110, 2024. 1, 4, 5, 7, 8

[41] Tristan Thrush, Ryan Jiang, Max Bartolo, Amanpreet Singh, Adina Williams, Douwe Kiela, and Candace Ross. Winoground: Probing vision and language models for visio-linguistic compositionality. In *IEEE/CVF Conference on Computer Vision and Pattern Recognition, CVPR*, pages 5228–5238, 2022. 2, 12

[42] Junyang Wang, Yuhang Wang, Guohai Xu, Jing Zhang, Yukai Gu, Haitao Jia, Ming Yan, Ji Zhang, and Jitao Sang. An llm-free multi-dimensional benchmark for mllms hallucination evaluation. *CoRR*, abs/2311.07397, 2023. 1, 4, 8

[43] Chenfei Wu, Shengming Yin, Weizhen Qi, Xiaodong Wang, Zecheng Tang, and Nan Duan. Visual chatgpt: Talking, drawing and editing with visual foundation models. *CoRR*, abs/2303.04671, 2023. 8

[44] Wenyi Xiao, Ziwei Huang, Leilei Gan, Wanggui He, Haoyuan Li, Zhelun Yu, Hao Jiang, Fei Wu, and Linchao Zhu. Detecting and mitigating hallucination in large vision language models via fine-grained AI feedback. *CoRR*, abs/2404.14233, 2024. 1, 2, 5, 7, 8, 16, 21

[45] Qinghao Ye, Haiyang Xu, Guohai Xu, Jiabo Ye, Ming Yan, Yiyang Zhou, Junyang Wang, Anwen Hu, Pengcheng Shi, Yaya Shi, Chenliang Li, Yuanhong Xu, Hehong Chen, Junfeng Tian, Qian Qi, Ji Zhang, and Fei Huang. mplug-owl: Modularization empowers large language models with multimodality. *CoRR*, abs/2304.14178, 2023. 1, 8

[46] Shukang Yin, Chaoyou Fu, Sirui Zhao, Tong Xu, Hao Wang, Dianbo Sui, Yunhang Shen, Ke Li, Xing Sun, and Enhong Chen. Woodpecker: Hallucination correction for multimodal large language models. *CoRR*, abs/2310.16045, 2023. 12

[47] Tianyu Yu, Yuan Yao, Haoye Zhang, Taiwen He, Yifeng Han, Ganqu Cui, Jinyi Hu, Zhiyuan Liu, Hai-Tao Zheng, Maosong Sun, et al. Rlhf-v: Towards trustworthy mllms via behavior alignment from fine-grained correctional human feedback. In *IEEE Conference on Computer Vision and Pattern Recognition, CVPR*, pages 13807–13816, 2024. 1, 4, 5, 7, 8, 16, 21

[48] Tianyu Yu, Haoye Zhang, Yuan Yao, Yunkai Dang, Da Chen, Xiaoman Lu, Ganqu Cui, Taiwen He, Zhiyuan Liu, Tat-Seng Chua, and Maosong Sun. RLAIF-V: aligning mllms through open-source AI feedback for super GPT-4V trustworthiness. *CoRR*, abs/2405.17220, 2024. 1, 4, 5, 8, 16, 21

[49] Zihao Yue, Liang Zhang, and Qin Jin. Less is more: Mitigating multimodal hallucination from an EOS decision perspective. In *Proceedings of the 62nd Annual Meeting of the Association for Computational Linguistics (Volume 1: Long Papers), ACL*, pages 11766–11781, 2024. 5, 8

[50] Tiancheng Zhao, Tianqi Zhang, Mingwei Zhu, Haozhan Shen, Kyusong Lee, Xiaopeng Lu, and Jianwei Yin. Vl-checklist: Evaluating pre-trained vision-language models with objects, attributes and relations. *CoRR*, abs/2207.00221, 2022. 2, 12

[51] Zhiyuan Zhao, Bin Wang, Linke Ouyang, Xiaoyi Dong, Jiaqi Wang, and Conghui He. Beyond hallucinations: Enhancing lvlms through hallucination-aware direct preference optimization. *CoRR*, abs/2311.16839, 2023. 1, 2, 4, 5, 8

[52] Kankan Zhou, Eason Lai, Wei Bin Au Yeong, Kyriakos Mouratidis, and Jing Jiang. ROME: evaluating pre-trained vision-language models on reasoning beyond visual common sense. In *Findings of the Association for Computational Linguistics: EMNLP*, pages 10185–10197, 2023. 12

[53] Yiyang Zhou, Chenhang Cui, Rafael Rafailov, Chelsea Finn, and Huaxiu Yao. Aligning modalities in vision large language models via preference fine-tuning. *CoRR*, abs/2402.11411, 2024. 1, 5, 8

[54] Yiyang Zhou, Chenhang Cui, Jaehong Yoon, Linjun Zhang, Zhun Deng, Chelsea Finn, Mohit Bansal, and Huaxiu Yao. Analyzing and mitigating object hallucination in large vision-language models. In *The Twelfth International Conference on Learning Representations, ICLR*, 2024. 5, 8

# Supplementary Materials

In Section A, we describe the choices of expert models and how expert models are executed when performing atomic tasks. The defined heuristics for [RELA]$_{spatial}$ and [ATTR]$_{spatial}$ are provided in Section B and Section C, respectively. Performance of error mitigation strategies based on MLLMs larger than 7B is shown in Section D. In Section E, we conduct a small scale of human evaluation for checking the performance of individual expert models. The ablation results with LLaVA-v1.5 as the base MLLM are illustrated in Section F. Limitations and potential future directions are elaborated in Section G. The performance of aligned MLLMs with DGPref on standard VQA settings is provided in Section H. Implementation details are provided in Section I. The visualization of responses from the MLLM before and after preference learning with DGPref is demonstrated in Section J and the visualization of our method extending to responses to other VQA types is available in Section K. Error cases in response decomposition are visualized in Section L. Human evalution settings are introduced in Section M. Lastly, prompt templates exploited are provided in Section N.

## A. Choices of Expert Models

**[DET]** detects and locates objects of interest in an image. To fulfill the goal, we exploit OWL-ViT [28], a strong open-vocabulary object detection model. Given the name of an object entity and an image, it returns the bounding boxes in the image containing instances of the object, together with a confidence score for each box.

**[RELA]** verifies whether a given relation exists between a subject and an object. As previous studies have shown, vision-language models are limited in spatial relation understanding [17, 41, 50]. Therefore we follow previous work [5, 39] to further separate spatial relations ([RELA]$_{spatial}$) from other general relations ([RELA]$_{general}$). For [RELA]$_{general}$, we convert the verification task to a binary VQA problem: "*Is the {sub} {relation} {obj}?*" and prompt BLIP-2 [46], a robust open-ended visual question answering (VQA) model, for an answer together with the image. If the answer is *yes*, we consider the relation to be depicted in the image, otherwise not. For [RELA]$_{spatial}$, we consider five categories of spatial relations (*left*, *right*, *top*, *bottom*, and *near*) as well as their synonyms (available in Appendix B.1). For the verification of these spatial relations, heuristic rules are applied over coordinates of [DET]($obj_s$) (i.e., detected bounding box(es) of the subject) and [DET]($obj_o$). Details of the heuristic rules are available in Appendix B.2.

**[ATTR]** is for the verification of an object attribute. As indicated in previous work [50, 52], vision-language models struggle at understanding object sizes, while frequently mentioned object sizes can be easily defined with heuristic rules based on the detected boxes of the objects. We divide the task into [ATTR]$_{general}$ and [ATTR]$_{size}$, according to the attribute to be verified. For [ATTR]$_{general}$, we convert the extracted attribute tuple into a binary question: "*Is the {obj} {attribute}?*" and prompt BLIP-2 [20] for an answer. Heuristic rules are applied for [ATTR]$_{size}$, when verifying five attributes about object sizes (*small*, *large*, *long*, *short*, *tall*), as well as their synonyms (available in Appendix C.1). The heuristic rules defined are provided in Appendix C.2.

**[COUNT]** checks if the number of detected bounding boxes for an object equals to the extracted number of counts from the response. No expert model is needed for this task.

**[OCR]** detects and reads text in an image. To achieve this goal, we leverage EasyOCR, a widely used OCR detection tool. It can read natural scene text and supports over 80 languages. During verification, it first detects all texts in the image and checks if there is an exact match between detected texts and the extracted mentioned image text fromt the response.

For [RELA]$_{general}$ and [ATTR]$_{general}$, there is a process of generating a binary question based on the extracted check-worthy content. This conversion may suffer from errors during response decomposition as well as the inflexibility of the manually defined template. Considering this, we included a grammar checker [30] to filter out questions likely to be grammatically incorrect. Given a question, the grammar checker will provide a score in the range 0 to 1, indicating how likely the question is to be fluent and grammatically correct. Questions with a low score will not be passed to expert models for verification.

The expert models are set empirically and a small scale human evaluation of individual experts on atomic tasks is provided in Appendix E. The results demonstrate that expert models perform well on these evaluation tasks. All expert models are replaceable. With the development of foundation models in a certain field (e.g., object detection), DecompGen could be potentially benefited. Meanwhile, DecompGen is always flexible to add other check-worthy aspects (e.g., checking commonsense mentioned in responses) as we can easily parts related to the aspect and insert expert models powerful regarding to the aspect using the proposed framework.

## B. Spatial Relations

The verification of some spatial relations, heuristic rules with object detection results (i.e., coordinates of detected objects) work well [5, 39]. We heuristically define the verification of five categories of spatial relations: *left*, *right*, *top*, *bottom* and *near*, as well as their synonyms.

## B.1. Synonyms of Spatial Relations

Below are synonyms included for each category of spatial relation:

**Left**: all relations including the word *left* (e.g., *to the left of* and *on the left of*).

**Right**: all relations including the word *right* (e.g., *to the right of* and *on the right of*).

**Top**: *above* and all relations including the word *top* (e.g., *on top of*).

**Bottom**: *below*, *under*, *beneath*, *underneath* and all relations including the word *bottom* (e.g., *in the bottom of*).

**Near**: *next*, *next to* and all relations including the word *near*.

---

**Algorithm 1** Verification for Spatial Relations

1: **procedure** $\text{VERIFY}_{\text{SP}}(\mathcal{B}_s, \mathcal{B}_o, rela)$
2:    **for** $b_s \in \mathcal{B}_s$ **do**
3:       **for** $b_o \in \mathcal{B}_o$ **do**
4:          **if** *rela* is *Left* **then**
5:             **if** $(x_{s,1}+x_{s,2} < x_{o,1}+x_{o,2})$ **then**
6:                **return** True
7:             **end if**
8:          **end if**
9:          **if** *rela* is *Right* **then**
10:             **if** $(x_{s,1}+x_{s,2} > x_{o,1}+x_{o,2})$ **then**
11:                **return** True
12:             **end if**
13:          **end if**
14:          **if** *rela* is *Top* **then**
15:             **if** $(y_{s,1}+y_{s,2} > y_{o,1}+y_{o,2})$ **then**
16:                **return** True
17:             **end if**
18:          **end if**
19:          **if** *rela* is *Bottom* **then**
20:             **if** $(y_{s,1}+y_{s,2} < y_{o,1}+y_{o,2})$ **then**
21:                **return** True
22:             **end if**
23:          **end if**
24:          **if** *rela* is *Near* **then**
25:             **if** $|x_{s,1}+x_{s,2} - x_{o,1}-x_{o,2}| <$ w * 0.1 **then**
26:                **return** True
27:             **end if**
28:             **if** $|y_{s,1}+y_{s,2} - y_{o,1}-y_{o,2}| <$ h * 0.1 **then**
29:                **return** True
30:             **end if**
31:          **end if**
32:       **end for**
33:    **end for**
34:    **return** False
35: **end procedure**

## B.2. Heuristic Rules for Spatial Relations

We provide the heuristic rules defined for the verification of spatial relations above, given the detected bounding boxes of the subject and the object: $\mathcal{B}_s$, $\mathcal{B}_o$. Each bounding box is represented as $\mathbf{b} = (x_1, y_1, x_2, y_2)$, where $x_1, y_1$ and $x_2, y_2$ are the coordinates for the bottom-left and top-right of a bounding box.

The detailed heuristic rules defined in the Algorithm 1, where h and w are for the height and width of images.

## C. Object Size Attributes

The verification of size-related attributes will be defined with heuristic rules, using the size of detected object bounding boxes. We consider five types of size-related attributes: *large*, *small*, *long*, *short* and *tall*, as well as their synonyms. Below are synonyms of these types attributes:

### C.1. Synonyms of Object Size Attributes

**Large**: *huge*, *big* and attributes including *large*.
**Small**: *tiny* and attributes including *small*.
**Long**: no other synonyms.
**Tall**: *high*.
**Short**: no other synonyms.

### C.2. Heuristic Rules for Object Size Attributes

We provide the heuristic rules defined for the verification of size-related attributes, given the detected bounding boxes of an object, $\mathcal{B}$. The detailed heuristic rules defined in the Algorithm 2.

## D. Performance of Error Mitigation Methods with Larger MLLMs

Due to the limitation of space, we only show performance of 7B models in Table 2. Here, we provide performance of error mitigation strategies applied on larger MLLMs. The results of these methods on the three evaluation benchmarks are provided in Table 5.

We are unable to conduct preference alignment based on 13B MLLMs as they cost more computational resources. However, we find that $\text{DGPref}_{\text{LLaVA}}$ and $\text{DGPref}_{\text{Qwen}}$, though only with 7B parameters, achieves competitive performance comparing with 12B or 13B baselines. Besides, we notice the most powerful aligned MLLMs (e.g., HSA-DPO and RLAIF-V), though have powerful performance in hallucination related metrics, suffer from **information loss**. Specifically, their COVER. (i.e., coverage rate) on AMBER is quite low, indicating their hallucination mitigation may be at the sacrifice of information loss. MLLMs with DGPref, in contrast, maintain the balance between error mitigation and informativeness the best.

| Model | Size | ObjHal | | MMHal | | AMBER | | | |
|---|---|---|---|---|---|---|---|---|---|
| | | CHAIR$_s$ ↓ | CHAIR$_i$ ↓ | Score ↑ | HalRate ↓ | s.CHAIR$_i$ ↓ | COVER. ↑ | HalRate ↓ | Cog. ↓ |
| HALVA | 13B | 45.4 | 12.8 | 2.58 | 0.45 | 6.4 | <u>52.6</u> | 30.4 | 3.2 |
| HSA-DPO | 13B | <u>5.3</u> | 3.2 | 2.61 | 0.48 | 2.1 | 47.3 | 13.4 | 1.2 |
| RLHF | 13B | 38.1 | 18.9 | 2.02 | 0.63 | 7.7 | 52.1 | 39.0 | 4.4 |
| RLHF-V | 13B | 12.2 | 7.5 | 2.81 | 0.49 | 6.3 | 46.1 | 25.1 | 2.1 |
| RLAIF-V | 12B | **3.3** | **1.8** | **3.36** | **0.29** | 5.4 | 46.4 | 27.1 | 1.1 |
| DGPref$_{LLaVA}$ | 7B | 10.3 | 2.6 | 2.59 | 0.36 | **1.2** | 51.2 | **7.8** | **0.5** |
| DGPref$_{Qwen}$ | 7B | 8.0 | <u>2.1</u> | <u>3.19</u> | <u>0.34</u> | <u>1.5</u> | **54.0** | <u>10.0</u> | <u>0.8</u> |

Table 5. Performance of existing works trying to mitigate errors in MLLMs' responses, with larger base MLLMs.

**Algorithm 2** Verification for Size-related Attributes

1: **procedure** VERIFY$_{SIZE}$($\mathcal{B}$, *attr*)
2:   **for** b ∈ $\mathcal{B}$ **do**
3:     **if** *attr* is *Large* **then**
4:       **if** ($\frac{y_2-y_1}{h} > 0.4 || \frac{x_2-x_1}{w} > 0.4$) **then**
5:         **return** True
6:       **end if**
7:     **end if**
8:     **if** *attr* is *Small* **then**
9:       **if** ($\frac{x_2-x_1}{w} < 0.3$ & $\frac{y_2-y_1}{h} < 0.3$ ) **then**
10:         **return** True
11:       **end if**
12:     **end if**
13:     **if** *attr* is *Long* **then**
14:       **if** ($\frac{y_2-y_1}{h} > 0.5 || \frac{x_2-x_1}{w} > 0.5$) **then**
15:         **return** True
16:       **end if**
17:     **end if**
18:     **if** *attr* is *Short* **then**
19:       **if** ($\frac{x_2-x_1}{w} < 0.3$ & $\frac{y_2-y_1}{h} < 0.3$ ) **then**
20:         **return** True
21:       **end if**
22:     **end if**
23:     **if** *attr* is *Tall* **then**
24:       **if** ($\frac{y_2-y_1}{h} > 0.4$) **then**
25:         **return** True
26:       **end if**
27:     **end if**
28:   **end for**
29:   **return** False
30: **end procedure**

| Expert | Accuracy |
|---|---|
| Object Detection | 86.8 |
| Spatial Relation | 88.5 |
| General Relation | 79.4 |
| OCR Text | 50.0 |
| Scale Attribute | 92.9 |
| General Attribute | 76.2 |

Table 6. Performance of individual experts on atomic verification tasks.

## E. Quality Check for Experts

To guarantee the quality of generated feedback, we need to ensure individual experts perform well on atomic reasoning tasks. In this section, we manually check the first 50 cases for expert performance. We define the accuracy of expert performance as number of correct predictions among all predictions. Here we do not consider errors from response decomposition nor errors from previous steps with other experts (i.e., the wrong detection result will lead to failure of relation verification, whereas, it will not be considered as an error from the relation expert). The performance of experts are shown in Table 6. The results demonstrate expert models have good performance on atomic verification tasks.

## F. Ablation Results with LLaVA-v1.5

In this section, we provide ablation results based on LLaVA-v1.5 [24] in the same setting as described in Section 3.3. The results are provided in Table 7.

From the results, we observe similar patterns to the ablation results based on Qwen-VL-Vhat [3]:
- DecompGen considering all five check-worthy aspects provides better feedback for preference dataset construction, than only considering object hallucinations.
- The performance of individual expert model matters to DecompGen. Poor expert performance leads to inaccurate DecompGen feedback, which will incur a low quality preference dataset.
- Using expert models for atomic verification tasks has a superiority over using human annotations. Human annotations are limited both in diversity and coverage.
- Compared with COCO images, VG images, with more

| Model | ObjHal | | MMHal | | AMBER | | | |
|---|---|---|---|---|---|---|---|---|
| | CHAIR$_s$ ↓ | CHAIR$_i$ ↓ | Score ↑ | HalRate ↓ | s.CHAIR$_i$ ↓ | COVER. ↑ | HalRate ↓ | Cog. ↓ |
| Base | 54.7 | 15.9 | 2.19 | 0.57 | 7.4 | 51.8 | 34.7 | 4.1 |
| Pref$_{Obj}$ | 13.3 | 2.8 | 2.25 | 0.45 | 2.0 | 43.8 | 8.4 | 0.4 |
| Pref$_{Obj-0.1}$ | 42.7 | 12.6 | 2.35 | 0.58 | 5.2 | 49.2 | 24.0 | 2.5 |
| Pref$_{Obj-GT}$ | 1.3 | 1.2 | 1.48 | 0.74 | 2.9 | 39.6 | 12.5 | 0.6 |
| DGPref$_{COCO}$ | 17.0 | 5.1 | 2.73 | 0.39 | 1.6 | 46.3 | 9.1 | 0.7 |
| DGPref$_{half}$ | 14.3 | 5.0 | 2.71 | 0.36 | 1.2 | 50.7 | 8.1 | 0.5 |
| DGPref | 10.3 | 2.6 | 2.59 | 0.36 | 1.2 | 51.2 | 7.8 | 0.5 |

Table 7. Ablation results of DecompGen for preference dataset construction, based on LLaVA-v1.5.

complicated visual information, can provide richer learning signals for MLLMs during preference learning.
- The amount of preference data matters to preference learning. It emphasizes the value of DecompGen for automatic preference data collection.

| Model | GQA | Text-VQA |
|---|---|---|
| LLaVA | 61.3 | 57.0 |
| DGPref$_{LLaVA}$ | 61.3 | 56.4 |

Table 8. Performance of LLaVA-v1.5 on VQA related dataset before (**LLaVA**) and after DPO training with DGPref (**DGPref$_{LLaVA}$**).

## G. Limitations

Our focus in this paper is on enhancing the trustworthiness of MLLMs by grounding their responses in visual context. Nonetheless, because of its modular architecture, DecompGen can also be adapted to incorporate assessments of other critical aspects, such as commonsense knowledge mentioned in responses. This flexibility allows DecompGen to generalize and collect preference data aimed at improving MLLMs' ability to generate accurate responses that integrate knowledge beyond image content.

Second, while DGPref is larger than human-annotated preference datasets, it is still considerably smaller than the pre-training datasets used for MLLMs. A promising future direction would be to apply our framework in the pre-training stage, enabling it to assess and filter high-quality data for MLLM pre-training.

Thirdly, DecompGen provides fine-grained and interpretable assessments. However, in preference learning, we currently aggregate feedback scores to classify responses as either preferred or rejected. This approach is due to limitations in existing preference learning algorithms, which rely on binary signals (i.e., preferred and rejected). With the development of preference optimization algorithms, De-compGen's detailed assessments could provide richer learning signals for more nuanced training.

| Dataset | Size |
|---|---|
| ObjHal | 300 |
| MMHal | 95 |
| AMBER | 1003 |

Table 9. Statistics of evaluation datasets

| Expert Model | # Parameters |
|---|---|
| OWL-ViT | 611M |
| EasyOCR | 27M |
| BLIP-2 | 3.4B |
| Grammar checker | 66M |
| Total | 4.1B |

Table 10. Model sizes of experts.

## H. Impact of Preference Alignment to Other Settings

We expect models to enhance trustworthiness after preference learning without sacrificing their helpfulness. We validate MLLMs' performance on standard VQA datasets [14, 38], after preference alignment, to demonstrate that our method will not have negative impact to their helpfulness. We choose LLaVA-v1.5 [24] as the base model as fine-tuning is essential for evaluation on these datasets and LLaVA-v1.5 has been fine-tuned on these datasets. The performance of the base MLLM and DGPref$_{LLaVA}$ is shown in Table 8. We observe DGPref$_{LLaVA}$ with enhanced trustworthiness achieves comparable performance to the base model on standard VQA datasets.

## I. Experiment Settings

**Dataset Statistics** We report the statistics of evaluation datasets in Table 9. For ObjHal, we follow [47, 48] to use the exact the same 300 instances sampled from the original ObjHal dataset for evaluation.

**Expert Models** We provide the expert model sizes in Table 10. For OWL-ViT, we set the detection threshold as 0.25. For the grammar checker model, we set the valid threshold as 0.75 (i.e., converted questions with a grammar chcker score below 0.75 will be filtered out).

**Package Versions** We implement all models under the PyTorch Library (version 2.4.0+cu121), with CUDA version 12.1. We use the NVIDIA H100 GPU, each with a dedicated memory of 80GB. For the implementation of LLaVA-v1.5 and Qwen-VL-Chat, we leverage the HuggingFace Library, with the llava-hf/llava-1.5-7b-hf and the Qwen/Qwen-VL-Chat checkpoints, respectively. The version of HuggingFace is 4.44.0. For the parameter-efficient tuning with LoRA, we leverage the package of PEFT, the version of which is 0.9.0.

For LLaVA-v1.5 7B, we train it with LoRA for two epochs, where the lora rank is 32, the lora alpha is 32, learning rate is 5e-6, the DPO hyper-parameter, $\beta$ is 0.1 and batch size is 8. We set the mini batch size as 4 and accumulate the gradients. It takes less than 7 hours with only one H100 GPU for DPO training. For Qwen-VL-Chat 7B, we train it with LoRA for one epochs, where the lora rank is 32, the lora alpha is 32, learning rate is 7e-6, the DPO hyper-parameter, $\beta$ is 0.1 and batch size is 8. We set the mini batch size as 1 and accumulate the gradients, taking about 5 hours with only one H100 GPU for DPO training. The models could be trained faster if using more than one H100 card. For the DGPref construction for Qwen-VL-Chat, we follow [44] to penalize object hallucinations twice.

For the evaluation of the MMHal dataset, GPT-4-0314, as it has been deprecated. We follow the Open AI document to use GPT-4-0613 alternatively.

## J. Visualization of Model Responses

In this section, we visualize the base model, LLaVA-v1.5 [24] and responses from DGPref$_{\text{LLaVA}}$. The comparison is shown in Table 11.

We observe the base model not only suffers from object hallucinations, but also provides inaccurate descriptions about attributes and relations. In contrast, DGPref$_{\text{LLaVA}}$ gives accurate responses without losing informativeness.

## K. Visualization of DecompGen Extending to Other VQA Type

In the paper, we focused on a specific VQA setting, generating detailed image descriptions, when constructing DGPref. We choose the setting because MLLMs are likely to give problematic response in this scenario [22] and this setting is also widely used for constructing preference datasets [21, 44, 47]. However, our method is not limited to this specific VQA setting. It can also be extended to other VQA question types.

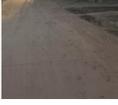

Figure 5. Annotation instructions provided to the annotators and an example from the questionnaire sent to annotators.

We query a base MLLM, LLaVA-v1.5 with questions and images from VQA-v2 [8], which covers diverse question types. Following the pipeline in Section 2, DecompGen generates feedback to model responses. The generated feedback is shown in Table 12. It highlights the flexibility of DecompGen and its potential to be scaled up for filtering out high-quality data for pre-training.

## L. Error Cases in Response Decomposition

In this section, we provide examples illustrating the error types of DecompGen during response decomposition, as described in Section 3.4. Specifically, we show three types of commonly seen errors of DecompGen during response decomposition: *invalid check-worthy parts extraction*, *incomplete extraction* and *hallucinations in extraction*. We give one example for each error type in Table 13.

## M. Human Evaluation

We asked four NLP researchers to participate in the human evaluation. Each two annotators are responsible for validating the same 15 examples. An example of the questionnaire distributed to annotators is shown in Figure 5. The agreements from annotators over the three metrics (i.e., fewer errors, more informative and preferable) are 0.13, 0.15 and 0.24 in terms of Cohen's Kappa $\kappa$, respectively. The agreements show that though annotators may be more subjective about which response has fewer errors or is more informative, they have a fair agreement on which response is more preferable, indicating the MLLM after preference learning with DGPref also aligns with human preferences to some extent.

| Image | Query | LLaVA Response | DGPref_LLaVA Response |
|---|---|---|---|
| 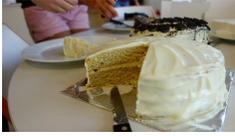 | Explain the narrative or story that the image seems to convey, detailing each part that contributes to it. | The image captures a delightful scene of people gathered around a table, enjoying a celebration or a special occasion, likely a birthday or an anniversary. There are two partially eaten cakes on the table, each accompanied by a knife. The cakes are visually enticing, and the knives signify that they have already been used to cut and serve the cakes to the guests. There is a person sitting near the cakes, possibly about to serve the cake or already enjoying the event. In addition to the cakes and knives, there are multiple forks and spoons on the table, suggesting that the guests might be using these utensils to enjoy the cake and other food items available at the gathering. A handbag is also present, indicating that this event may be attended by friends or family members who have brought personal belongings with them. | The image depicts a scene where a large cake is placed on a table, and it is being sliced. There are multiple cakes in the scene, with one being the main focus, and others are present in the background. A person is reaching for a cake, possibly to slice it or to enjoy it. The presence of multiple cakes and the act of slicing one of them suggest a celebration or gathering, where people are enjoying desserts. The person's action of reaching for the cake indicates that they are actively participating in the event and are likely excited to share the cake with others. The overall atmosphere of the image conveys a sense of joy and togetherness. |
| 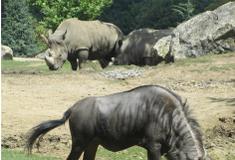 | Out of the four animals, how many giraffes are there in the image? | There are two giraffes in the image. | There are no giraffes in the image. |
| 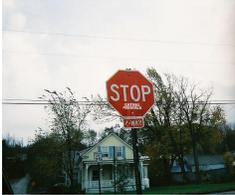 | Provide a thorough description of the given picture. | The image features a stop sign standing on a street corner. The stop sign is positioned near a house, with a street sign located below it. The stop sign stands out with its bright red color, making it easily noticeable to drivers. In addition to the stop sign, there are multiple cars parked or driving by in the scene. Some cars are parked closer to the house, while others are located further away, on the street. Overall, it's a typical city street scene with a stop sign directing traffic. | A stop sign is in the foreground, with a house in the background. The stop sign is large and is positioned in front of a house. The sky is cloudy, and there are trees in the background. The house is situated in a neighborhood, and the stop sign is a prominent feature in the scene. The house is situated in the foreground, and the stop sign is positioned in front of it. |

Table 11. Visualization of generated response between LLaVA-v1.5 and DGPref_LLaVA. We highlight the errors in the base model responses with red and mark the corresponding correct parts in DGPref_LLaVA's responses with blue.

| Image | Question | Response | Feedback |
|---|---|---|---|
| 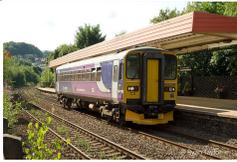 | Is this train at the station? | Yes, the train is at the station, sitting on the tracks. | `Existence:` train✓, station✓, tracks✓<br>`Attributes`$_{general}$: (sitting, train)✓<br>`Relations`$_{general}$: (train, on, tracks)✓ |
| 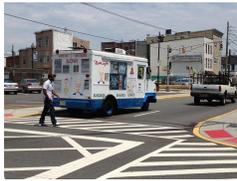 | What is the man doing in the street? | The man is walking across the street in front of a blue and white ice cream truck. | `Existence:` ice cream truck✓, man✓, street✓<br>`Attributes`$_{general}$: (blue and white, ice cream truck)✓, (walking, man)✓<br>`Relations`$_{general}$: (man, across, street)✓, (man, in front of, ice cream truck)✓ |
| 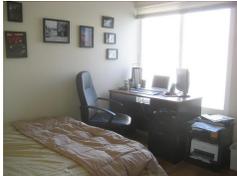 | Is there a mirror in the room? | Yes, there is a mirror in the room, located above the desk. | `Existence:` desk✓, mirror✗ |
| 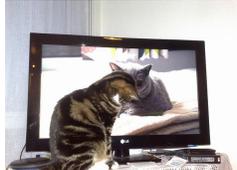 | How many cats are in the image? | There are two cats in the image. | `Existence:` cat✓<br>`Counts:` (2, cat)✗ |

Table 12. Visualization of DecompGen extended to other VQA question types. Skipped verifications are not shown in the table.

| Response | Decomposition Result |
|---|---|
| The image depicts a busy city street filled with a variety of vehicles, including cars and trucks. The street is lined with streetlights, and there are multiple traffic lights positioned at different locations, ensuring the smooth flow of traffic. Several cars can be seen in the scene, with one car in the foreground and others further back. A truck is visible on the left side of the street. In the midst of the busy traffic, there is a cow statue standing on the street, adding an interesting element to the urban landscape. | `Existence`: street, traffic light, cow statue, car, truck, location, vehicle, streetlight<br>`Attributes`: (busy, street), (streetlights, street), (car, street), (truck, street), (cow, statue)<br>`Relations`$_{general}$: (car, on, street), (traffic light, at location), (cow statue, on, street)<br>`Relations`$_{spatial}$: (truck, left, street) |
| The image features two large airplanes parked next to each other on an airport tarmac, likely waiting for passengers or for maintenance. There is a small truck situated near the right side of the planes, possibly for maintenance or support purposes. In addition to the airplanes and the truck, there are several cars and trucks scattered around the scene, indicating possible airport traffic or service vehicles. Two people can be spotted near the middle of the image, possibly airport staff or travelers. | `Existence`: airplane, tarmac, truck, car, person, airport, planes<br>`Count`: (two, airplane)<br>`Attributes`$_{size}$: (large, airplane)<br>`Relations`$_{general}$: (car, scattered around, scene), (truck, scattered around, scene)<br>`Relations`$_{spatial}$: (airplane, near, airplane), (truck, right, planes), (person, near, image) |
| No, it is not going to rain. | `Existence`: person, rain |

Table 13. Visualization of errors of DecompGen during response decomposition. The first row shows extracting invalid check-worthy aspect, the second row shows extracting incomplete parts and the last row is an example of hallucination during decomposition. The error part is highlighted with red.

> Provide a thorough description of the given image.
>
> What is this photo about? Please answer in great detail.
>
> Provide a thorough description of the given picture.
>
> Explain the narrative or story that the image seems to convey, detailing each part that contributes to it.
>
> Please provide a detailed description of the image. Describe the visual elements, colors, shapes, textures, and any objects or people present along with the overall mood or atmosphere portrayed in the image.
>
> Please provide a detailed description of the image, including its visual elements, such as colors, shapes, textures, objects, and people.
>
> Provide an intricate description of the image, capturing its visual elements, including colors, shapes, textures, objects, and any people present.
>
> Compose a detailed account of the image, encompassing its visual characteristics, like colors, shapes, textures, objects, and any human subjects, by paying careful attention to the specifics.

Figure 6. Eight diverse instructions for generating detailed image descriptions.

## N. Prompt Templates

In this section, we provide the exact prompting templates used in the paper.

**Diverse Templates for Detailed Image Descriptions:** The diverse templates for detailed image descriptions are used in preference data collection (Section 2.3) and the evaluation on ObjHal (Section 3.1). We use the same diverse templates as [44, 47, 48].

**Templates for the Decomposition of Responses:** In Section 2.1, we mentioned that we leverage the in-context learning capability of LLMs for extracting check-worthy parts. Specifically, we provide eight in-context examples for extracting each check-worthy aspect from a response. For *object existence*, the template with examples is available in Figure 7; For *object relations*, the template is available in Figure 8; For *object attributes*, the template is available in Figure 9; For *counts of objects*, the template is available in Figure 10; For *image texts*, the template is available in Figure 11.

You are a expert writer. Given a textual description ([DESP]), your goal is to extract the visible entities ([ENT]) in description. Entities should include objects, people, animals and extracted entities are separated with commas. Some examples are provided below:

[DESP]: The image features a woman standing in a kitchen, preparing food. She is smiling and seems to be enjoying the cooking process. The kitchen is well-equipped with various appliances such as ovens, a microwave, and a sink. On the counter, there's a bowl and an assortment of items, including a cup, a few bottles, a couple of bowls, and a spoon. A chair is also visible in the scene. The woman is wearing a white shirt and blue jeans, adding a casual and friendly touch to the image
[ENT]: (woman, kitchen, food, oven, microwave, sink, bowl, cup, bottle, bowl, spoon, chair, shirt, jeans)

[DESP]: The image displays a baseball game in progress, with a batter taking a swing at an incoming pitch. The batter's baseball bat is close to the ground, as he is in the process of making contact with the ball. The scene is dynamic and action-packed. In addition to the batter, there are several other players on the field, including two teammates close to the batter, and three more players in the background. Some players are positioned near the pitcher, while others are scattered across the field. Other baseball equipment can be spotted in the scene, such as a baseball glove and
[ENT]: (batter, baseball bat, ball, player, pitcher, baseball glove)

[DESP]: The image features a large pan with a single slice of pizza, loaded with toppings such as olives, mushrooms, and cheese. The pizza slice is placed in the middle of the pan, occupying a significant portion of the serving area. The close-up view of the pizza slice emphasizes its delicious toppings and overall presentation.
[ENT]: (pan, pizza, olives, mushroom, cheese)

[DESP]: The image displays a tasty-looking sandwich placed in a Styrofoam container, accompanied by a side of sauce in a separate plastic container. The sandwich is cut in half, making it easy to eat and enjoy. There are two sandwiches in the scene, with one slightly larger than the other. Various carrots can be seen in the image, including some that are placed on and near the sandwiches. The carrots are scattered around the container, adding a vibrant touch to the scene.
[ENT]: (sandwich, container, sauce, carrot)

[DESP]: The image portrays a city street with a black SUV parked along the sidewalk. There are a few people in the scene, including a man standing near the parked car, and two other individuals further away from the car. The street is surrounded by trees, providing a pleasant atmosphere. There are also some traffic signs visible, indicating that the area might be a designated parking zone. The overall scene depicts a typical urban setting with parked vehicles and pedestrians going about their day.
[ENT]: (city street, SUV, sidewalk, people, man, tree, traffic sign, vehicle, pedestrian)

[DESP]: In the image, a man is standing outside a large clock tower, seemingly posing or admiring the impressive structure. The tower features a massive clock face with Roman numerals, which is visible in the foreground. The man is positioned on the left side of the clock tower. Additionally, there is another clock located further down the tower, although it is not as prominent as the main clock face. The man appears to be the primary focus of the scene, with the clock tower serving as a backdrop for the picture.
[ENT]: (man, clock tower, Roman numeral, clock)

[DESP]: The image showcases a beautiful aerial view of a large lake and an island in the middle of the water. The airplane wing is in the foreground, covering most of the image and occupying the left side. The wing's angle and positioning provide a sense of perspective, emphasizing the vastness of the lake and island. The clear sky and sunlight highlight the picturesque scene.
[ENT]: (lake, island, water, airplane, wing's angle, sky, sun light)

[DESP]: The image features a scene of a person riding a horse in a body of water, likely a river. The horse and rider are the main focus of the scene. The water is flowing beneath them, providing a unique setting for the activity. The horse seems to be enjoying the water as the rider sits on its back, ready to continue their journey. The rider is wearing a helmet, ensuring their safety while crossing the body of water.
[ENT]: (person, horse, water, river, rider, helmet)

Please extract [ENT] of this [DESP]: {*description*}

Figure 7. The prompt template for extracting object entities in a response.

> You are a expert writer. Given a textual description ([DESP]), your goal is to extract the described spatial relations between objects ([RELA]) in the format of triplets (subject, relation, object). The relations cover spatial relations and actions. For instance, a cat to the left of a dog can be represented as (cat, to the left of, dog); a man holding an umbrella can be represented as (man, holding, umbrella). Make sure there is no adjectives for either subjects or objects. If no relations involved, please respond NONE. Some examples are provided below:
>
> [DESP]: The image showcases a kitchen counter with a variety of oranges and a juicer in the foreground. There are several oranges arranged on the counter in different positions, with some closer to the front and others further back. The juicer is prominently placed on the counter, ready to be used to extract fresh orange juice. In addition to the juicer and oranges, a bottle can be seen placed near the edge of the counter.
> [RELA]: (orange, on, counter); (bottle, near the edge of, counter)
>
> [DESP]: The image shows a red and white train traveling on a set of train tracks. The train is located near the center of the scene and is quite prominent as it moves along the tracks. There are a couple of trucks visible in the image, one on the left side and another on the right side of the frame. Two traffic lights can also be seen - one close to the truck on the left and the other near the left edge of the image. Additionally, a person is present in the scene, standing near the center-left area of the image
> .[RELA]: (train, on, train tracks); (truck, on left side of, frame); (truck, on right side of, frame); (traffic light, close to, truck); (traffic light, near the left edge of, frame); (person, center-left of, image)
>
> [DESP]: The image portrays a group of people riding horses along a path, with some of them wearing hats. In total, there are at least five people and six horses visible on the road. The horses are spread out across the scene, with some closer to the left side, others near the center, and a few more towards the right side. The riders are in various positions, some in the foreground, and others further back, all enjoying their horseback riding experience together.
> [RELA]: (horse, on, road); (horse, left side of, image); (horse, center of, image); (horse, right of, image); (people, riding, horse); (people, wearing, hat)
>
> [DESP]: The image showcases a man in a red jacket and grey pants standing in a snow-covered area. He is holding ski poles and has skis on, ready for skiing. The man appears to be enjoying his time on the slopes. The scene seems to be captured during a winter day, with the man being the main focus.
> [RELA]: (man, holding, ski poles)
>
> [DESP]: The image displays a dining table with a variety of food items arranged on it. There are three plate trays filled with different types of food, including sandwiches, vegetables, and some other dishes. The sandwiches are spread across the plates, with one on the left side, one in the middle, and another on the far right. In addition to the sandwiches, there are multiple carrots, with two on the left side, one in the middle, and another on the far right corner. There are also two bottles on the table, one towards the upper left corner
> [RELA]: (food, on, dining table); (sandwiches, left of, plate); (sandwiches, in the middle of, plate); (sandwiches, right of, plate); (bottle, on, table); (bottle, left corner of, table)
>
> [DESP]: The image features a cat standing on a tiled floor in a bathroom. The cat is looking down, possibly intrigued by an object on the floor. The bathroom has a sink nearby, and the floor is adorned with a mosaic tile pattern. The cat appears to be relaxed in the bathroom environment.
> [RELA]: (cat, on, floor)
>
> [DESP]: The image displays a blue and gold military airplane parked on a runway. It is the main focus of the scene, with its impressive size and color scheme. There is a person standing near the back of the airplane, possibly a member of the air force or an technician. In the background, there is another airplane, which appears to be slightly smaller and located further away. A truck can be seen on the right side of the image, possibly used for maintenance or support purposes. The overall setting suggests an active military airfield or base.
> [RELA]: (person, back of, airplane); (truck, right of, image)
>
> [DESP]: The image shows a red and white train traveling on a set of train tracks. The train is located near the center of the scene and is quite prominent as it moves along the tracks. There are a couple of trucks visible in the image, one on the left side and another on the right side of the frame. Two traffic lights can also be seen - one close to the truck on the left and the other near the left edge of the image. Additionally, a person is present in the scene, standing near the center-left area of the image.
> [RELA]: (train, on, train tracks); (truck, on left side of, frame); (truck, on right side of, frame); (traffic light, close to, truck); (traffic light, near the left edge of, frame); (person, center-left of, image)
>
> Please extract relations ([RELA]) of this [DESP]: {*description*}

Figure 8. The prompt template for extracting object relations in a response.

You are a expert writer. Given a textual description ([DESP]), your goal is to extract the described visible attributes of objects ([ATTR]) in the format of tuples (attribute, object). Attributes can include colors, materials, shapes, activities, features and scales. Please make sure the attribute is visible (avoid attributes like: beautiful, delicious, colorful, comfortable, warm, etc). If no attributes extracted, please respond NONE. Some examples are provided below:

[DESP]: The image features a shirtless man standing at the edge of the water, holding a surfboard. He appears to be wearing a tight wetsuit, and the surfboard he holds is prominently visible. The beach is bathed in warm sunlight, creating a pleasant atmosphere for the surfer.
[ATTR]: (shirtless, man); (standing, man); (tight, wetsuit)

[DESP]: The image shows a dining table with a white plate in the center, containing a delicious pepperoni pizza. The pizza is ready to be eaten, and it is accompanied by a variety of food items. There is a portion of broccoli placed near the pizza, along with a bowl of veggies, adding a nutritious element to the meal. Additionally, there is a cup positioned on the table, possibly containing a beverage like soda. The table setting is complemented by a chair placed on one side, offering a comfortable spot to enjoy the meal.
[ATTR]: (white, plate)

[DESP]: The image features two blue plastic lawn chairs on a sidewalk, placed under a large umbrella. The chairs are placed close together, and the umbrella provides shade from above. Both chairs are accompanied by bags, likely containing personal items or belongings. The sidewalk is adorned with potted plants and small trees, creating a pleasant atmosphere for relaxation. There are five potted plants visible in the image, some placed closer to the chairs and others extending further down the sidewalk.
[ATTR]: (blue, lawn chair); (plastic, lawn chair); (large, umbrella); (small, tree)

[DESP]: A dog is sitting on top of a wooden boat, looking out over the water, likely enjoying the view or waiting for its owner to return. The boat is moored in a harbor, with several other boats of varying sizes in the background. These other boats are situated near the main boat where the dog is sitting, making the scene lively and active.
[ATTR]: (wooden, boat); (sitting, dog); (moored, boat)

[DESP]: The image displays a tasty-looking sandwich placed in a Styrofoam container, accompanied by a side of sauce in a separate plastic container. The sandwich is cut in half, making it easy to eat and enjoy. There are two sandwiches in the scene, with one slightly larger than the other. Various carrots can be seen in the image, including some that are placed on and near the sandwiches. The carrots are scattered around the container, adding a vibrant touch to the scene.
[ATTR]: NONE

[DESP]: The image depicts a busy city intersection with a large white bus driving down the street. The street is characterized by tall buildings on both sides, and there are numerous people walking around the area. A yellow taxi cab is also present in the scene, sharing the busy street with the bus. Several pedestrians can be seen throughout the image, with some walking close to each other, and others spread out at varying distances. A handbag is noticeable with one of the pedestrians, adding to the bustling atmosphere of the city.
[ATTR]: (large, bus); (white, bus); (tall, building); (yellow, taxi cab)

[DESP]: The image features a woman standing on a beach with her surfboard. She is wearing a pink and black wetsuit, and her surfboard is placed by her side. She has her hand on her head, possibly as a means of checking the waves or preparing for her surfing session. The beach setting appears to be well-suited for water sports like surfing.
[ATTR]: (pink and black, wetsuit)

[DESP]: The image features a large gray elephant walking in a dirt area alongside a rock wall. The elephant seems to be exploring its surroundings. In addition to the elephant, there is a bird perched in the top right corner of the scene. The area has a mix of dirt, rocks, and a few plants. The overall setting appears to be a natural habitat for the elephant, providing a habitat-like environment for the animal.
[ATTR]: (large, elephant); (gray, elephant); (walking, elephant); (rock, wall)

Please extract [ATTR] of this [DESP]: {*description*}

Figure 9. The prompt template for extracting object attributes in a response.

You are a expert writer. Given a textual description ([DESP]), your goal is to extract the counting of objects ([COUNT]), in the format of tuples (number, OBJ). If the number is one, there is no need for extraction. Avoid implicit counting, like several, a few, a group of, etc. If no counting involved in the description, please respond NONE. Some examples are provided below:

[DESP]: The image features two blue plastic lawn chairs on a sidewalk, placed under a large umbrella. The chairs are placed close together, and the umbrella provides shade from above. Both chairs are accompanied by bags, likely containing personal items or belongings. The sidewalk is adorned with potted plants and small trees, creating a pleasant atmosphere for relaxation. There are five potted plants visible in the image, some placed closer to the chairs and others extending further down the sidewalk.
[COUNT]: (five, potted plants)

[DESP]: The image captures a beautiful night scene in a city, featuring a large building with a clock tower. The building has a prominent clock on its side, which stands out against the dark sky. The clock tower is part of a building with columns, giving it an impressive architectural look. In addition to the clock tower, there are two more clocks visible in the scene. A bench can be seen in the foreground, providing a place for people to sit and enjoy the view. The night sky above the cityscape gives the scene an atmospheric ambiance.
[COUNT]: (three, clocks)

[DESP]: The image features a hospital room containing two beds, with one of them being a cot. The beds are positioned next to each other. The room also has a chair placed nearby, close to the beds. Additionally, there are four pillows on the beds, providing comfort to the patients. There are two people in the room, one standing close to the left bed and the other near the right bed. The room appears to be organized and ready for use, with all necessary furniture and accessories in place.
[COUNT]: (two, beds); (four, pillow); (two, people)

[DESP]: The image features a large public transit bus driving down a city street, possibly in a park setting. The bus is situated in the middle of the scene, with a traffic light visible close to the bus. There are several cars on the street, with one on the left side of the bus and three more on the right side. In addition to the vehicles, there is a person walking along the sidewalk, and another individual closer to the right side of the scene. A fire hydrant can be seen near the bus and another one further down the street, adding to the urban setting.
[COUNT]: (two, fire hydrant)

[DESP]: The image depicts a horse grazing on grass in a field, surrounded by a tall fence. The horse is positioned towards the center of the scene, and it appears to be enjoying the fresh green grass. The fence is visible on the left and right sides of the horse, as well as on the top, creating a secure enclosure for the animal.
[COUNT]: None

[DESP]: The image displays an open and well-lit indoor space. At the center of the scene, a dining table is set with a variety of objects. There are multiple chairs surrounding the table, with one on each side and one in the foreground. Additionally, there is a chair in the background and another one in the middle of the room. Among the items placed on the table are a wine glass and a bowl, suggesting that it is set up for a meal or gathering. The scene also features several bottles, including one on the left side of the table and two more on the right
[COUNT]: (three, bottle)

[DESP]: The image is a group picture of people gathered together on a beach. They are sitting around a dining table, which is placed in the middle of the scene. Various items are scattered across the table, including bottles, cups, and a knife. There are several chairs placed around the table, as well as a few umbrellas set up nearby. Some people are sitting on chairs, while others are standing around the table. The atmosphere appears to be joyful and social, with people spending time together on a beautiful beach day.
[COUNT]: NONE

[DESP]: In the image, there are two police officers riding on the back of horses. They are positioned in front of a large building and seem to be walking down a road. The officers appear to be patrolling the street or providing a visible presence for the community. The scene also includes a bus driving behind the officers on the left side. Additionally, there are two cars visible in the image: one car is located at the far right side of the scene, and the other car is near the right edge of the photo. A person can also be seen standing near the right edge of the image, possibly observing the officers
[COUNT]: (two, police officer); (two, car)

Please extract [COUNT] of this [DESP]: {*description*}

Figure 10. The prompt template for extracting the counts of an object in a response.

You are a expert writer. Given a textual description ([DESP]), your goal is to extract the mentioned scene text (text in the image) in the description with [TEXT]. Different pieces of scene texts are separated with commas. If no scene text mentioned, please respond NONE. Some examples are provided below:

[DESP]: The image depicts a large green military truck with a camouflage paint, parked in front of a church. Several other trucks are parked around the area, indicating a gathering or event. In total, there are eight trucks visible in the scene, with the main military truck being the largest and most prominent. Some people can be seen in the vicinity of the parked trucks, suggesting that they might be attending the event or assisting in some way.
[TEXT]: NONE

[DESP]: The image depicts a modern shopping center with sleek, glass-fronted stores lining a spacious, polished stone walkway. Shoppers stroll by, carrying colorful bags, while a few people relax on benches under decorative planters filled with vibrant flowers. The center is bright, illuminated by natural light streaming in from large skylights above. On the left side of the image, a prominent store entrance features a large sign above the doorway. The sign reads Luxury Fashion in elegant, gold letters against a black background.
[TEXT]: (Luxury Fashion)

[DESP]: The image features a large giraffe standing in an open field with a mix of bushes and trees nearby. The giraffe appears to be looking out over the field, possibly searching for food or observing its surroundings. The giraffe's long neck stands out prominently, highlighting its unique features. The field is spacious, providing ample room for the giraffe to roam and interact with its environment.
[TEXT]: NONE

[DESP]: The image showcases a small blue and white boat anchored on a large lake. The boat is positioned towards the center of the lake, and it appears to be a family-sized boat with a cabin. There are two people visible on the deck of the boat, possibly enjoying some time on the water. In the surrounding area, there are a few other boats, with one located closer to the left side and another towards the right. Moreover, there is a car parked on the shoreline, close to the left edge of the image.
[TEXT]: NONE

[DESP]: The image shows a lively New York City street corner. A green street sign in the foreground reads "Broadway" in bold white letters, with another sign below it indicating "W 42nd St." A large, illuminated billboard on a nearby building advertises a Broadway show, with the text "Phantom of the Opera" in elegant, golden script. On the sidewalk, a small kiosk displays a colorful poster that says NYC Tours and Tickets $25 in red and blue. A yellow taxi passes by, with an ad on its roof that reads "Visit Central Park." The scene is bustling with people and traffic.
[TEXT]: (Broadway; W 42nd St; Phantom of the Opera; NYC Tours; Tickets $25; Visit Central Park.)

[DESP]: The image captures a serene coastal scene at sunset. The vast expanse of the calm sea stretches out to the horizon, where the sky is painted in soft hues of orange, pink, and purple. Gentle waves lap against a sandy shoreline, where a few scattered seashells and small rocks are visible. The beach is mostly empty, with only a few distant seagulls flying low over the water. On the right side of the image, a weathered wooden sign is partially embedded in the sand. The sign, painted in faded white letters, reads "Private Beach - No Trespassing." The text stands out against the rustic wood, adding a subtle human element to the otherwise untouched natural scene. The overall atmosphere is tranquil and inviting, with the focus drawn to the quiet message on the sign.
[TEXT]: (Private Beach - No Trespassing.)

[DESP]: In the image, a young woman is riding a skateboard with her hands in front of her. She is wearing a helmet and knee pads, providing her with safety while performing the activity. The scene takes place in a park-like setting, with greenery surrounding the area. The young skateboarder is enjoying her time outdoors, showcasing her skill and confidence.
[TEXT]: NONE

[DESP]: The image captures a classic London street on a cloudy day, with historic buildings lining the cobblestone road. Red double-decker buses and black cabs move along the street, while pedestrians, some with umbrellas, walk along the sidewalks. The architecture is a mix of old brick facades and modern glass structures, blending the city's rich history with its contemporary vibe. A traditional black iron lamp post features a decorative sign attached to it, reading Baker Street in bold, white letters on a dark green background. The scene exudes a sense of charm and history, with the "Baker Street" sign anchoring the image in its famous setting.
[TEXT]: (Baker Street)

Please extract [TEXT] of this [DESP]: {*description*}

Figure 11. The prompt template for extracting mentioned image texts in a response.